\documentclass{article} 
\usepackage{iclr2027_conference,times}
\iclrfinalcopy

\usepackage{amsmath,amsfonts,bm}

\def\eqref#1{equation~\ref{#1}}

\def\1{\bm{1}}

\DeclareMathAlphabet{\mathsfit}{\encodingdefault}{\sfdefault}{m}{sl}
\SetMathAlphabet{\mathsfit}{bold}{\encodingdefault}{\sfdefault}{bx}{n}

\usepackage{algorithm}
\usepackage{algorithmic}
\usepackage{xspace}
\usepackage{xcolor}
\usepackage{color}
\usepackage{dsfont}
\usepackage{subcaption}
\usepackage{amsfonts}       
\usepackage{amsmath}

\usepackage{hyperref}
\usepackage{url}
\usepackage{float}
\usepackage[normalem]{ulem}
\usepackage{enumitem}
\usepackage{amsthm}
\usepackage{algorithm}
\usepackage{algorithmic}
\usepackage{graphicx}
\usepackage{booktabs}
\usepackage{multirow}
\usepackage{multicol}
\usepackage{colortbl}
\usepackage{arydshln}
\usepackage{amssymb}

\usepackage{wrapfig}
\usepackage{etoc}
\etocdepthtag.toc{mtchapter}
\etocsettagdepth{mtchapter}{subsection}
\etocsettagdepth{mtappendix}{none}

\usepackage[most]{tcolorbox}
\usepackage{csquotes}
\usepackage{anyfontsize}
\usepackage{comment}
\usepackage{float}
\usepackage{cuted}
\usepackage{tikz}
\usepackage{listings,multicol}
\usepackage[
    framemethod=tikz,
    skipbelow=\topskip,
    skipabove=\topskip
]{mdframed}
\newtcolorbox[
    list inside=prompt,
    auto counter,
    number within=section
]{promptbox}[1][]{%
    colback=white,
    colframe=black!60,
    colbacktitle=black!60,
    coltitle=white,
    fontupper=\normalsize,
    boxsep=5pt,
    left=0pt,
    right=0pt,
    top=0pt,
    bottom=0pt,
    boxrule=1pt,
    #1
}

\newcommand{\our}{SkillCome\xspace}
\title{SkillCome: Group Contrast Skill Optimization with Dual Memory}

\author{Haolin Li$^{\spadesuit,\clubsuit}$\ \ ,\  Feng Hong$^{\diamondsuit}$, Ang Li$^{\clubsuit}$, Chilin Fu$^{\clubsuit}$, Weichang Wu$^{\clubsuit}$, \\ \textbf{Ya Zhang}$^{\diamondsuit}$, \textbf{Yanfeng Wang}$^{\diamondsuit}$, \textbf{Xiaolu Zhang}$^{\clubsuit,\dagger}$, \textbf{Jiangchao Yao}$^{\diamondsuit}$\thanks{Corresponding Authors} \\
  $^{\spadesuit}$Fudan University \, $^{\diamondsuit}$Shanghai Jiao Tong University \, $^{\clubsuit}$Ant Group
} 

\begin{document}

\setcounter{footnote}{1}
\maketitle
\lhead{Preprint}

\begin{abstract}
Skill evolution improves the capabilities of large language models by analyzing trajectories generated under a given skill and modifying the skill accordingly.
Existing approaches typically generate a single trajectory per question.
However, this provides insufficient optimization signals since it requires inferring effective skill edits from a solitary path.
It is difficult to pinpoint which actions caused the failure in a failed trajectory, or to determine which actions in a successful one should be incorporated into the skill.
Furthermore, they rely on a local batch of trajectories for analysis, making the optimization direction susceptible to noisy evidence.
To address these, we propose SkillCome, a \underline{Skill}-evolution method based on group \underline{C}ontrast \underline{o}ptimization with dual \underline{me}mory.
For each question, SkillCome generates $n$ trajectories and performs group contrast analysis to precisely identify key behavioral divergences between successful and failed trajectories, offering reliable optimization signals.
The dual memory system further accumulates evidence from historical steps to track patterns shared across different groups, leading to more generalized optimization directions.
Together, SkillCome builds a systematic optimization process that transforms experience from observed successful trajectories into reusable skills.
Extensive experiments on six benchmarks spanning question answering, reasoning, and agentic tasks demonstrate the effectiveness of our method.
SkillCome consistently outperforms baselines across five models of varying families and scales, with gains up to \textbf{+5.69\%} points.
Our code is available at \url{https://github.com/tdlhl/SkillCome}.

\end{abstract}

\begin{figure}[H]
    \centering
    \includegraphics[width=\linewidth]{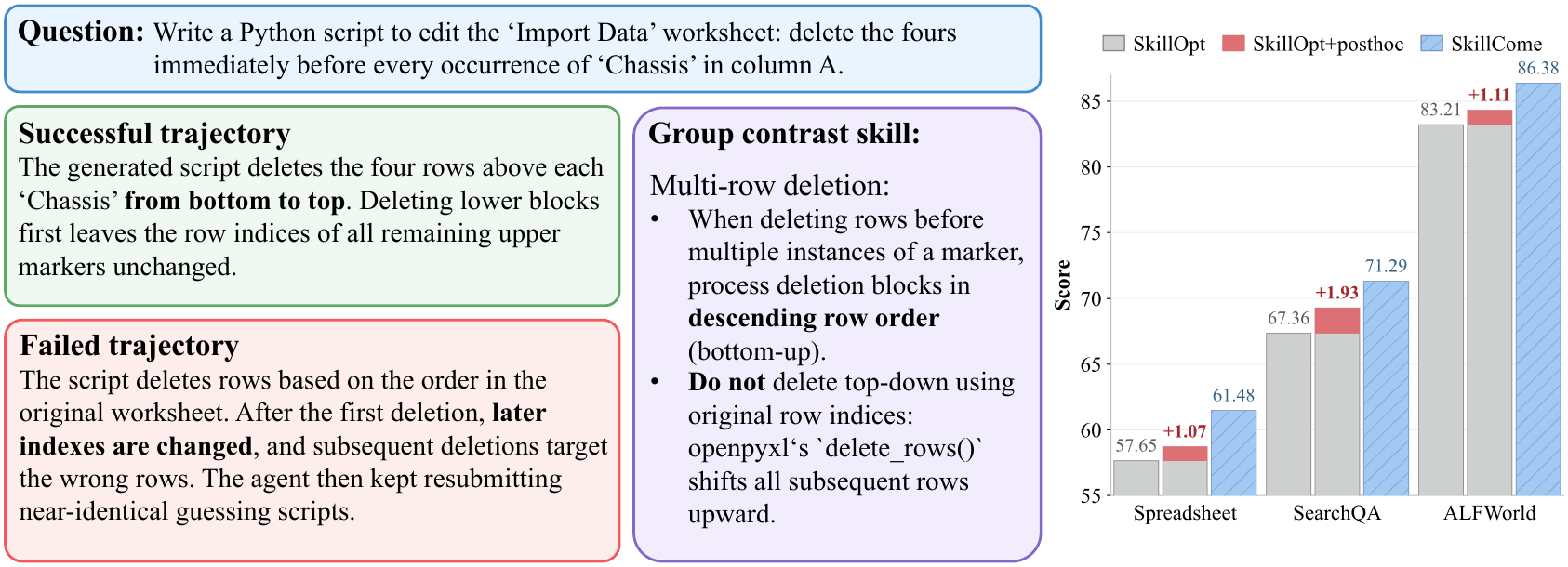}
    \caption{
    \textbf{Left}: An example of a skill learned by \our on SpreadsheetBench.
    Through group contrast analysis, \our learns to delete rows in descending order to avoid index shifts invalidating subsequent deletions, a skill absent from the single-rollout methods.
    \textbf{Right}: Post-hoc experiment on Qwen3.6-35B-A3B.
    After SkillOpt training, we perform one round of group contrast analysis.
    This single post-hoc skill edit can already improve performance on three datasets.
    }
    \label{fig:intro}
    \vspace{-1.0em}
\end{figure}

\section{Introduction}

Large language models (LLMs) can be improved through parametric optimization of model weights or non-parametric optimization of external context~\citep{ouyang2022training,wang2023voyager}. Skills~\citep{xu2026agent,li2026skillsbench}, typically derived from human experience or LLM-generated trajectories~\citep{jiang2026sok,Bai2026SkillDAGST}, provide an effective form of non-parametric optimization.
Recent works on skill evolution further optimize skills iteratively~\citep{Wang2025ReinforcementLF,Lin2026MUSEAutoskillSA}.
Specifically, a target model executes tasks under the current skill, and an optimizer model analyzes the resulting trajectories to propose skill revisions, which are then evaluated on validation sets~\citep{zhang2026coevoskills,Zhu2026SkillCoachSR}.

Despite its effectiveness, skill evolution requires more than an iterative loop of generating and evaluating edits.
A sound optimization process requires two key components: a reliable optimization signal that indicates how the current solution should be updated, and an optimizer to organize the signals and guide update directions.
Parametric optimization has developed sophisticated mechanisms for both training objectives~\citep{wei2022finetuned,rafailov2023direct} and optimizer states~\citep{kingma2015adam,loshchilov2018decoupled}.
This motivates us to examine both the construction of optimization signals and the design of optimizer state for skill evolution.

At the optimization signal level, the outcome of a trajectory does not directly reveal how the skill should change.
Existing approaches typically analyze one trajectory per question~\citep{alzubi2026evoskill,Wu2026CoEvolvingLD}, leaving it unclear which actions caused failure or contributed to success.
At the optimizer level, current methods primarily observe trajectories from a local batch, which can provide noisy or incomplete evidence.
A valuable skill pattern can appear in trajectories scattered across different steps.
Consequently, a single batch lacks sufficient data to substantiate this recurring pattern, leaving the optimizer to overlook it for inclusion in the skill.

These limitations call for an optimization process with reliable learning signals and a stateful optimizer.
In parametric optimization, group relative rewards among responses to the same question have been widely adopted as optimization signals~\citep{shao2024deepseekmath}.
For skill evolution, comparing successful and failed trajectories within a group similarly reveals behavioral divergences associated with opposite outcomes.
The optimizer can then derive reliable optimization signals from observed successful attempts, rather than inferring a potential solution solely from a failed trajectory.
Figure~\ref{fig:intro} illustrates how same-question contrast reveals a row-deletion error in worksheet-operation tasks and yields a practical skill.
We further examine this idea by performing a single post-hoc group contrast skill revision using existing SkillOpt trajectories for the same question.
As shown on the right of Figure~\ref{fig:intro}, this already improves performance without additional training rollouts.
For such evidence to guide optimization direction, a memory system is necessary to preserve historical observations.
Analogous to momentum in gradient descent~\citep{pmlr-v28-sutskever13}, the memory can retain patterns learned from previous steps.
The optimizer can then combine current findings with accumulated information, continually calibrating its update direction as further evidence becomes available.
Such a stateful optimization process helps turn scattered evidence into generalizable skills, preventing noise in individual steps from dominating the optimization process.

In this work, we propose \our, a skill-evolution method that provides more reliable optimization signals while guiding optimization directions with historical evidence.
Our method reshapes trajectory generation by introducing grouped rollouts, collecting multiple trajectories for each question.
We incorporate within-group contrast between successful and failed trajectories to conventional reflection.
These comparisons help identify key behavioral differences and precisely transform trajectory evidence into targeted skill revisions, improving optimization reliability and data efficiency.
We further introduce a dual-memory system that connects current observations with historical information.
Contrast memory preserves contrast patterns learned between successful and failed trajectories within the same group, while failure memory retains failure patterns shared by different groups.
By continuously interacting with the dual memory, the optimizer can discover shared patterns across questions and use them to refine the skill update directions.
The two synergistic components establish a comprehensive optimization process for skill evolution.
Experiments across diverse model-benchmark pairs demonstrate that \our can discover effective skills that are difficult to learn for single-rollout methods, leading to improved skill evolution and performance.
In summary, our contributions are threefold:
\begin{itemize}
    \item We introduce group rollouts and within-group contrast optimization under the same skill.
    This provides reliable optimization signals by pinpointing the key divergences between successful and failed trajectories, grounding skill revisions in observed experience.
    \item To calibrate the direction of skill optimization, we propose a dual-memory system storing historical evidence.
    The system recognizes recurring patterns from different steps and builds support for generalizable skills, preventing noisy updates from individual batches.
    \item Extensive experiments on six benchmarks validate the effectiveness of our method across five models with varying scales. \our enables effective skill evolution and delivers consistent improvements, with average gains up to \textbf{19.55\%} compared to no skill.
\end{itemize}

\section{Related Work}

\paragraph{From Skills to Skill Evolution.}

Prompt-based methods~\citep{wei2022chain,NEURIPS2023_271db992,pryzant2023automatic} have demonstrated that LLMs can be improved by simply refining the textual instructions~\citep{wang2023selfconsistency,liu2023pre}.
Beyond general prompts, agent skills further include task-specific experience, including tool-use strategies and relevant references~\citep{Zheng2025SkillWeaverWA,shi2026skill1}.
They organize experience derived from human expertise or agent trajectories into natural language, optionally accompanied by executable scripts and supporting resources~\citep{shen2026skillfoundry}.
By making such experience editable, skills can also incorporate new lessons from execution feedback~\citep{yang2026autoskill}.
This has motivated various research on skill evolution, where agent trajectories are analyzed to iteratively revise skills and improve subsequent task performance~\citep{liu2026skillforge,ouyang2026skillos}.
They explicitly optimize skill artifacts through repeated generation and revision~\citep{liu2026skillforge,ouyang2026skillos}.
Trace2Skill~\citep{ni2026trace2skill} extracts local lessons from execution traces in parallel and consolidates them into transferable procedural guidance.
SkillOpt~\citep{yang2026skillopt} develops a structured optimization procedure that converts trajectories into bounded skill edits, with edit aggregation and validation-based acceptance.
SkillOpt-Lite~\citep{shen2026skillopt} and OEO~\citep{liu2026rethinking} delegate parts of the evolution process to an agent rather than a human-implemented pipeline, allowing the agent to automatically organize the evolution.
These studies establish generated trajectories as a basis for skill evolution, with different designs for extracting experience and learning edits.
While our method focuses on how to better construct the optimization process based on grouped rollouts.
\section{Method}

\begin{figure*}[tbp]
    \centering
    \includegraphics[width=\linewidth]{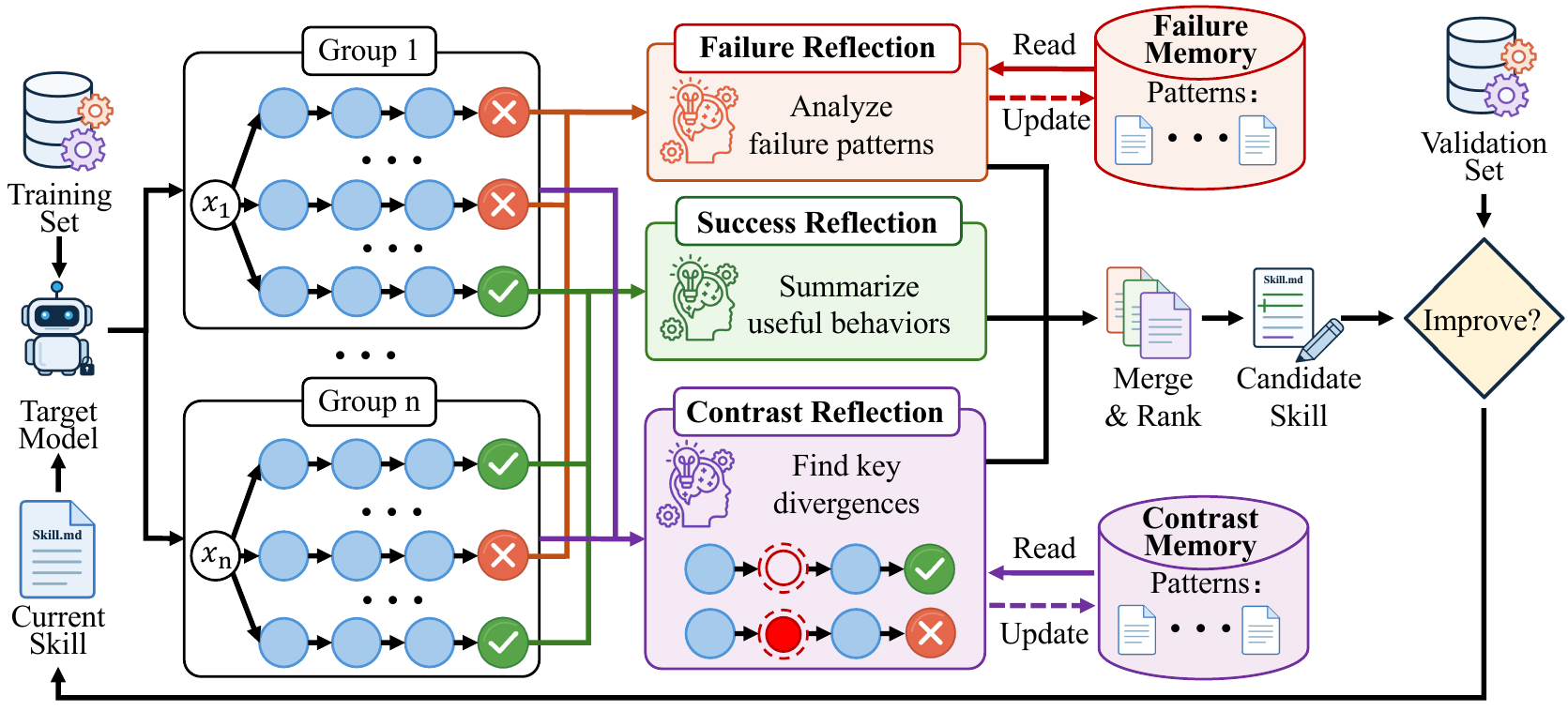}
    \vspace{-1.0em}
    \caption{ 
    Overview of \our. We first formulate the training as grouped rollouts.
    The generated trajectories are routed to either ``Success Reflection'' or ``Failure Reflection'' based on their outcome correctness.
    Groups with mixed outcomes undergo ``Contrast Reflection''.
    During reflection, the optimizer model reads and updates failure memory and contrast memory to find generalizable optimization directions.
    The updated skill is accepted if it improves the validation score.
    }
    \label{fig:method}
    \vspace{-0.5em}
\end{figure*}

In this section, we first formulate the skill-evolution problem in \S~\ref{sec:preliminary}.
We then describe grouped rollouts in \S~\ref{sec:grouped_rollout}.
\S~\ref{sec:group_contrast} presents our group contrast skill optimization procedure.
The overall framework of \our is illustrated in Figure~\ref{fig:method}.

\subsection{Problem Formulation}
\label{sec:preliminary}

In skill evolution, an external skill document is optimized to improve the performance of LLM agents on given tasks.
A skill $s$ consists of a natural-language document together with any accompanying executable scripts.
The document is incorporated into the agent context during execution.

We distinguish two model roles: the target model $M_{\mathrm{tar}}$ executes tasks under the current skill, while the optimizer model $M_{\mathrm{opt}}$ analyzes the resulting trajectories and proposes skill revisions.
The two roles can be the same or different LLMs, whose parameters remain frozen throughout skill evolution.
Given an input task $x$ and a skill document $s$, a trajectory is sampled as $\tau \sim M_{\mathrm{tar}}(\cdot \mid x,s)$.
The trajectory $\tau$ records model outputs, tool interactions when applicable, and the final answer.
A task-specific evaluator assigns a binary score $R(x,\tau) \in \{0,1\}$ according to the task's success criterion.
Given a dataset $\mathcal{D}$, the objective is to find a skill that maximizes expected performance:
\begin{equation}
    \max_s \;
    \sum_{x \in \mathcal{D}}
    \mathbb{E}_{\tau \sim M_{\mathrm{tar}}(\cdot \mid x,s)}
    \left[R(x,\tau)\right].
    \label{eq:skill_objective}
\end{equation}
In practice, we use three disjoint data splits: $\mathcal{D}_{\mathrm{train}}$ for collecting training trajectories, $\mathcal{D}_{\mathrm{val}}$ for evaluating the updated skills, and $\mathcal{D}_{\mathrm{test}}$ for testing the final skill.
Starting from an initial skill $s^0$, the target model executes training tasks under the current skill $s^t$ at optimization step $t$.
$M_{\mathrm{opt}}$ analyzes these trajectories and their evaluation feedback, then proposes and selects edits that add, remove, or replace content from $s^t$.
Applying the selected edits yields a new candidate skill $s_{\mathrm{cand}}^t$.

The candidate is evaluated on the validation split using the same target model and execution environment.
We denote the measured validation score of a skill $s$ by:
\begin{equation}
V(s) =
\frac{1}{|\mathcal{D}_{\mathrm{val}}|}
\sum_{x \in \mathcal{D}_{\mathrm{val}}}
R\!\left(x,\tau_x(s)\right),
\label{eq:validation_score}
\end{equation}
where $\tau_x(s)$ denotes the validation trajectory generated for task $x$ using skill $s$.
The candidate skill $s_{\mathrm{cand}}^t$ is accepted only if its score exceeds the recorded validation score of the previous skill:
\begin{equation}
    s^{t+1} =
    \begin{cases}
        s_{\mathrm{cand}}^t,
        & V\!\left(s_{\mathrm{cand}}^t\right) > V\!\left(s^t\right), \\
        s^t,
        & \text{otherwise}.
    \end{cases}
    \label{eq:skill_acceptance}
\end{equation}
After iteratively running the training-updating-validation loop, we retain the skill with the highest validation score as $s^{\star}$ and evaluate it on $\mathcal{D}_{\mathrm{test}}$.


\subsection{Grouped Rollout}
\label{sec:grouped_rollout}
To obtain comparable trajectory evidence for each question, we organize the training process in the form of group rollouts.
At optimization step $t$, we sample $m$ distinct training questions $\{x_i^t\}_{i=1}^{m} \subseteq \mathcal{D}_{\mathrm{train}}$.
For each question, the target model $M_{\mathrm{tar}}$ independently generates $n$ trajectories under the same current skill $s^t$:
\begin{equation}
    \tau_{ij}^t
    \sim M_{\mathrm{tar}}\!\left(\cdot \mid x_i^t,s^t\right),
    \qquad
    i=1,\ldots,m,
    \quad
    j=1,\ldots,n,
    \label{eq:grouped_sampling}
\end{equation}
where the subscripts $i$ and $j$ are the question index and its rollout index, respectively.
The skill remains fixed throughout the current optimization step.
Thus, the trajectories within a group represent alternative executions of the same question.
Each trajectory receives a score $r_{ij}^t = R(x_i^t,\tau_{ij}^t)$.
We organize the scored trajectories for each question into a group $\mathcal{G}_i^t$ and denote the collection of all groups in the batch by $\mathcal{G}^t$:
\begin{equation}
    \mathcal{G}_i^t =
    \left\{
        \left(\tau_{ij}^t,r_{ij}^t\right)
    \right\}_{j=1}^{n},
    \qquad
    \mathcal{G}^t =
    \left\{\mathcal{G}_i^t\right\}_{i=1}^{m}.
    \label{eq:trajectory_groups}
\end{equation}
The batch contains $m$ groups and $B=mn$ rollouts in total.

For each group, let $\mathcal{G}_i^{t,+}$ and $\mathcal{G}_i^{t,-}$ denote the subsets of successful trajectories with $r_{ij}^t=1$ and failed trajectories with $r_{ij}^t=0$, respectively.
Across the batch, we collect these trajectories into success and failure pools:
\begin{equation}
    \mathcal{T}^{t,+} =
    \bigcup_{i=1}^{m} \mathcal{G}_i^{t,+},
    \qquad
    \mathcal{T}^{t,-} =
    \bigcup_{i=1}^{m} \mathcal{G}_i^{t,-}.
    \label{eq:reflection_pools}
\end{equation}
These pools provide evidence for success reflection and failure reflection, respectively.
And we collect groups containing both successful and failed trajectories into the mixed-group subset:
\begin{equation}
    \mathcal{G}_{\mathrm{mix}}^t =
    \left\{
        \mathcal{G}_i^t \in \mathcal{G}^t
        \;\middle|\;
        \mathcal{G}_i^{t,+} \neq \varnothing
        \;\text{and}\;
        \mathcal{G}_i^{t,-} \neq \varnothing
    \right\}.
    \label{eq:mixed_groups}
\end{equation}
Each mixed group is passed to the subsequent group contrast reflection, where successful trajectories serve as concrete references for examining failures and guiding skill edits.

\subsection{Group Contrast Skill Optimization}
\label{sec:group_contrast}

\subsubsection{Dual Memory System}
\label{sec:dual_memory}

Since the optimizer $M_{\mathrm{opt}}$ does not have an infinite context length, each optimization step can analyze only a limited set of trajectories.
Skill edits derived from one step may overfit to a particular question, while evidence supporting a generalizable pattern may be distributed across multiple steps.
To connect these observations, we maintain a failure memory $\mathcal{M}_{\mathrm{f}}$ and a contrast memory $\mathcal{M}_{\mathrm{c}}$, both initialized as empty.
$\mathcal{M}_{\mathrm{f}}$ preserves failure patterns extracted through conventional reflection.
$\mathcal{M}_{\mathrm{c}}$ contains contrastive patterns learned from the behavioral differences between successful and failed trajectories within the same group.
We keep $\mathcal{M}_{\mathrm{f}}$ and $\mathcal{M}_{\mathrm{c}}$ separate to distinguish failure-only patterns from contrast patterns that provide more reliable optimization signals grounded in observed successful alternatives.

Rather than storing lengthy historical trajectories, each memory consists of concise patterns:
\begin{equation}
    \mathcal{M} =
    \left\{
        (p_k,u_k)
    \right\}_{k=1}^{K},
    \label{eq:memory_patterns}
\end{equation}
where $K$ is the number of learned patterns, $p_k$ is the natural-language description of the $k$-th pattern, and $u_k$ denotes the set of distinct training groups supporting it.
The size of $u_k$ indicates how many different questions support the pattern, rather than how many trajectories.
Patterns supported by more questions provide stronger evidence for broadly applicable skill edits.

During trajectory analysis, $M_{\mathrm{opt}}$ interacts with the corresponding memory.
It compares the current trajectories with the memory and proposes memory updates alongside its skill edits.
An update may introduce a new pattern, refine an existing description, or associate additional supporting groups with an existing pattern without changing its description.
For the latter two cases, the support set is updated as $u_k \leftarrow u_k \cup u_{\mathrm{new}}$, where $u_{\mathrm{new}}$ contains the newly associated group identifiers.
We denote the proposed memory updates by $\Delta\mathcal{M}$, which may be empty when no update is needed.
This read-and-update process allows a pattern initially observed in one step to accumulate evidence from different groups across optimization steps, without requiring additional training rollouts.

\subsubsection{Conventional Failure and Success Reflection}
\label{sec:conventional_reflection}

At optimization step $t$, the trajectory pools $\mathcal{T}^{t,-}$ and $\mathcal{T}^{t,+}$ obtained from \S\ref{sec:grouped_rollout} are analyzed separately by $M_{\mathrm{opt}}$ to propose skill edits.
Failure reflection examines what went wrong and how the skill could prevent similar failures, whereas success reflection extracts useful practices worth retaining.
Taking failure reflection as an example, the optimizer uses the current skill, failed trajectories, and failure memory to generate candidate skill edits and optional memory updates:
\begin{equation}
    \left(
        \mathcal{P}_{\mathrm{f}}^t,
        \Delta\mathcal{M}_{\mathrm{f}}^t
    \right)
    \leftarrow
    M_{\mathrm{opt}}\!\left(
        s^t,
        {\mathcal{T}}^{t,-},
        \mathcal{M}_{\mathrm{f}}
    \right),
    \label{eq:failure_reflection}
\end{equation}
where $\mathcal{P}_{\mathrm{f}}^t$ contains the proposed skill edits.
Historical failure patterns from $\mathcal{M}_{\mathrm{f}}$ help $M_{\mathrm{opt}}$ relate current mistakes to problems in previous steps and formulate edits supported by different questions.
Success reflection is performed through separate optimizer calls and similarly produces candidate edits $\mathcal{P}_{\mathrm{s}}^t$.
Note that trajectories from the mixed groups still contribute to conventional reflection, while they are additionally examined through group contrast reflection described next.

\subsubsection{Group Contrast Reflection}
\label{sec:group_contrast_reflection}

To accurately determine what to correct and what to retain in the skill, we introduce the group contrast reflection.
It compares successful and failed trajectories within the same group to track critical divergences that account for their opposite outcomes.
For each mixed group $\mathcal{G}_i^t \in \mathcal{G}_{\mathrm{mix}}^t$, a separate optimizer call generates candidate edits and optional updates to the contrast memory:
\begin{equation}
    \left(
        \mathcal{P}_{\mathrm{c},i}^t,
        \Delta\mathcal{M}_{\mathrm{c},i}^t
    \right)
    \leftarrow
    M_{\mathrm{opt}}\!\left(
        s^t,
        \mathcal{G}_i^{t,+},
        \mathcal{G}_i^{t,-},
        \mathcal{M}_{\mathrm{c}}
    \right).
    \label{eq:group_contrast_reflection}
\end{equation}
Here, $M_{\mathrm{opt}}$ is instructed to first locate critical divergences between the successful and failed trajectories.
It then examines which actions in the successful trajectories were omitted or executed incorrectly in the failures, and relates these differences to missing or insufficiently actionable guidance in $s^t$.
Successful trajectories thus provide concrete references that the target model has actually done, offering reliable optimization signals for skill editing.

The contrast memory $\mathcal{M}_{\mathrm{c}}$ further connects these same-question comparisons with patterns observed in previous groups.
A new comparison can then strengthen or refine an existing pattern instead of being interpreted as an isolated case.
This guides skill optimization toward generalizable directions supported by multiple questions, rather than fitting a single case.
And the proposed memory updates are integrated after each analysis, making them visible to subsequent steps.

\subsubsection{Evidence-Guided Skill Revision}
\label{sec:evidence_guided_revision}

We collect the contrast edits as $\mathcal{P}_{\mathrm{c}}^t = \bigcup_{\mathcal{G}_i^t \in \mathcal{G}_{\mathrm{mix}}^t} \mathcal{P}_{\mathrm{c},i}^t$ and combine them with the failure and success edits.
Through dedicated merge calls of the optimizer $M_{\mathrm{opt}}$, we consolidate overlapping edits and resolves conflicting suggestions:
\begin{equation}
    \mathcal{P}^t =
    \operatorname{Merge}_{M_{\mathrm{opt}}}\!\left(
        \mathcal{P}_{\mathrm{f}}^t
        \cup \mathcal{P}_{\mathrm{s}}^t
        \cup \mathcal{P}_{\mathrm{c}}^t
    \right).
    \label{eq:merge_proposals}
\end{equation}

When the number of edits in $\mathcal{P}^t$ exceeds the given edit budget $L^t$, $M_{\mathrm{opt}}$ will select at most $L^t$ edits.
Ranking prioritizes contrast edits and failure edits over success edits.
It also accounts for the number of supporting groups to find more generalizable skill edits.

Based on the above procedure, the optimizer integrates complementary evidence:
Conventional reflection captures shared difficulties and potentially useful practices; group contrast reflection provides practical guidance grounded in grouped rollouts.
In addition, the dual memory system further extends the supporting evidence across steps.
Applying the selected final edits $\Delta^t \subseteq \mathcal{P}^t$ produces the candidate skill:
\begin{equation}
    s_{\mathrm{cand}}^t =
    \operatorname{Apply}(s^t,\Delta^t),
    \qquad
    |\Delta^t| \leq L^t.
    \label{eq:apply_group_edits}
\end{equation}


The candidate skill $s_{\mathrm{cand}}^t$ is evaluated and accepted according to Eq.~\ref{eq:skill_acceptance}, after which the target model $M_{\mathrm{tar}}$ collects the next batch of grouped rollouts under the resulting skill.
Memory updates are retained even when a candidate is rejected.
Since an unsuccessful edit does not invalidate the underlying observations, which may be supported by different groups of rollouts in future steps.
Across iterations, group contrast deepens the analysis of individual questions and provides accurate optimization signals.
While the dual-memory system plays a momentum-like role to regulate optimization direction by retaining the patterns observed beyond the current batch.
For final evaluation, we test the target model on $\mathcal{D}_{\mathrm{test}}$ using the selected best skill $s^\star$.

\section{Experiments}

\providecommand{\na}{--}
\definecolor{ssoblue}{RGB}{220,232,255}
\definecolor{rowgray}{RGB}{238,238,238}

\providecommand{\dvalp}[2]{}
\renewcommand{\dvalp}[2]{%
  \hphantom{\textsubscript{\tiny\ensuremath{+#2}}}%
  #1\textsubscript{\tiny\textcolor{green!50!black}{\ensuremath{+#2}}}%
}

\providecommand{\dvaln}[2]{}
\renewcommand{\dvaln}[2]{%
  \hphantom{\textsubscript{\tiny\ensuremath{-#2}}}%
  #1\textsubscript{\tiny\textcolor{red!70!black}{\ensuremath{-#2}}}%
}

\begin{table*}[t]
\centering
\caption{
Performance across six benchmarks and five models.
``\na'' denotes unsupported VQA evaluation for text-only LLMs.
Subscripts in \textcolor{green!40!black!80!white}{green}/\textcolor{red!70!black!80!white}{red} represent the change relative to ``No skill''.
The best and second-best scores within each model are in \textbf{bold} and \uline{underlined}, respectively.
}
\label{tab:skill_results}
\vspace{-5pt}
\setlength{\tabcolsep}{1.5pt}
\renewcommand{\arraystretch}{1.10}
\resizebox{\textwidth}{!}{%
\begin{tabular}{llccccccc}
\toprule
\textbf{Model} & \textbf{Skill} & \textbf{Spreadsheet} & \textbf{SearchQA} & \textbf{LiveMath} & \textbf{ALFWorld} & \textbf{OfficeQA} & \textbf{DocVQA} & \textbf{Avg} \\
\midrule

\multirow{6}{*}{\shortstack[l]{Qwen3.8\\Flash-Next}}
& No skill & 48.58 & 67.50 & 33.73 & 81.53 & 23.48 & 90.37 & 57.53 \\
\cmidrule(lr){2-9}
& Init skill & \dvaln{45.55}{3.03} & \dvalp{67.97}{0.47} & \dvalp{35.14}{1.41} & \dvalp{82.47}{0.94} & \dvalp{30.75}{7.27} & \dvalp{91.38}{1.01} & \dvalp{58.88}{1.35} \\
& Trace2Skill & \dvalp{69.66}{21.08} & \dvalp{69.93}{2.43} & \dvaln{31.60}{2.13} & \dvalp{85.07}{3.54} & \dvalp{49.83}{26.35} & \dvalp{91.31}{0.94} & \dvalp{66.23}{8.70} \\
& SkillOpt-Lite & \dvalp{\uline{73.75}}{25.17} & \dvalp{72.43}{4.93} & \dvalp{41.75}{8.02} & \dvalp{\uline{86.19}}{4.66} & \dvalp{\uline{60.30}}{36.82} & \dvalp{\uline{91.71}}{1.34} & \dvalp{71.02}{13.49} \\
& SkillOpt & \dvalp{72.51}{23.93} & \dvalp{\uline{72.57}}{5.07} & \dvalp{\uline{49.53}}{15.80} & \dvalp{84.70}{3.17} & \dvalp{59.80}{36.32} & \dvalp{91.24}{0.87} & \dvalp{\uline{71.73}}{14.20} \\
& \our & \dvalp{\textbf{75.71}}{27.13} & \dvalp{\textbf{74.14}}{6.64} & \dvalp{\textbf{63.68}}{29.95} & \dvalp{\textbf{91.79}}{10.26} & \dvalp{\textbf{63.68}}{40.20} & \dvalp{\textbf{92.78}}{2.41} & \dvalp{\textbf{76.96}}{19.43} \\
\midrule
\multirow{6}{*}{\shortstack[l]{DeepSeek-V4\\Pro-0813}}
& No skill & 47.78 & 66.57 & 28.07 & 66.98 & 53.71 & \na & 52.62 \\
\cmidrule(lr){2-9}
& Init skill & \dvalp{48.40}{0.62} & \dvalp{67.07}{0.50} & \dvalp{31.37}{3.30} & \dvalp{79.10}{12.12} & \dvalp{54.56}{0.85} & \na & \dvalp{56.10}{3.48} \\
& Trace2Skill & \dvalp{54.98}{7.20} & \dvalp{69.21}{2.64} & \dvalp{33.25}{5.18} & \dvalp{80.04}{13.06} & \dvaln{46.45}{7.26} & \na & \dvalp{56.79}{4.16} \\
& SkillOpt-Lite & \dvalp{\uline{59.08}}{11.30} & \dvalp{69.29}{2.72} & \dvalp{32.08}{4.01} & \dvalp{80.60}{13.62} & \dvalp{54.73}{1.02} & \na & \dvalp{59.16}{6.53} \\
& SkillOpt & \dvalp{57.38}{9.60} & \dvalp{\uline{70.86}}{4.29} & \dvalp{\uline{48.35}}{20.28} & \dvalp{\uline{84.32}}{17.34} & \dvalp{\uline{56.76}}{3.05} & \na & \dvalp{\uline{63.53}}{10.91} \\
& \our & \dvalp{\textbf{62.10}}{14.32} & \dvalp{\textbf{72.50}}{5.93} & \dvalp{\textbf{53.30}}{25.23} & \dvalp{\textbf{89.18}}{22.20} & \dvalp{\textbf{59.29}}{5.58} & \na & \dvalp{\textbf{67.27}}{14.65} \\
\midrule
\multirow{6}{*}{GLM-5.3}
& No skill & 45.73 & 67.93 & 36.08 & 86.94 & 20.78 & \na & 51.49 \\
\cmidrule(lr){2-9}
& Init skill & \dvalp{52.44}{6.71} & \dvalp{68.57}{0.64} & \dvalp{37.03}{0.95} & \dvalp{91.23}{4.29} & \dvalp{27.84}{7.06} & \na & \dvalp{55.42}{3.93} \\
& Trace2Skill & \dvaln{43.15}{2.58} & \dvalp{70.50}{2.57} & \dvalp{36.79}{0.71} & \dvalp{88.81}{1.87} & \dvalp{27.03}{6.25} & \na & \dvalp{53.26}{1.76} \\
& SkillOpt-Lite & \dvalp{\uline{68.23}}{22.50} & \dvalp{\textbf{77.43}}{9.50} & \dvalp{\uline{65.09}}{29.01} & \dvalp{88.43}{1.49} & \dvalp{\uline{46.62}}{25.84} & \na & \dvalp{\uline{69.16}}{17.67} \\
& SkillOpt & \dvalp{66.90}{21.17} & \dvalp{74.29}{6.36} & \dvalp{45.05}{8.97} & \dvalp{\uline{92.35}}{5.41} & \dvalp{46.28}{25.50} & \na & \dvalp{64.97}{13.48} \\
& \our & \dvalp{\textbf{71.53}}{25.80} & \dvalp{\uline{77.07}}{9.14} & \dvalp{\textbf{67.92}}{31.84} & \dvalp{\textbf{95.90}}{8.96} & \dvalp{\textbf{48.65}}{27.87} & \na & \dvalp{\textbf{72.21}}{20.72} \\
\midrule
\multirow{7}{*}{\shortstack[l]{DeepSeek-V4\\Flash}}
& No skill & 47.33 & 64.85 & 31.60 & 59.51 & 5.41 & \na & 41.74 \\
\cmidrule(lr){2-9}
& Init skill & \dvalp{48.58}{1.25} & \dvalp{65.43}{0.58} & \dvalp{35.38}{3.78} & \dvaln{56.91}{2.60} & \dvalp{11.66}{6.25} & \na & \dvalp{43.59}{1.85} \\
& Trace2Skill & \dvalp{51.07}{3.74} & \dvalp{71.50}{6.65} & \dvalp{33.25}{1.65} & \dvalp{\uline{79.10}}{19.59} & \dvalp{12.84}{7.43} & \na & \dvalp{49.55}{7.81} \\
& SkillOpt-Lite & \dvalp{57.38}{10.05} & \dvalp{68.00}{3.15} & \dvalp{33.73}{2.13} & \dvalp{61.01}{1.50} & \dvalp{24.66}{19.25} & \na & \dvalp{48.96}{7.22} \\
& SkillOpt & \dvalp{\uline{59.16}}{11.83} & \dvalp{\uline{71.57}}{6.72} & \dvalp{\uline{68.63}}{37.03} & \dvalp{74.44}{14.93} & \dvalp{\uline{26.01}}{20.60} & \na & \dvalp{\uline{59.96}}{18.22} \\
& \our & \dvalp{\textbf{62.54}}{15.21} & \dvalp{\textbf{73.36}}{8.51} & \dvalp{\textbf{74.29}}{42.69} & \dvalp{\textbf{84.14}}{24.63} & \dvalp{\textbf{50.51}}{45.10} & \na & \dvalp{\textbf{68.97}}{27.23} \\
\midrule
\multirow{7}{*}{\shortstack[l]{Qwen3.6\\35B-A3B}}
& No skill & 41.99 & 66.00 & 26.65 & 66.04 & 37.16 & 89.77 & 54.60 \\
\cmidrule(lr){2-9}
& Init skill & \dvaln{40.75}{1.24} & \dvalp{66.23}{0.23} & \dvalp{30.19}{3.54} & \dvalp{79.48}{13.44} & \dvalp{\uline{38.68}}{1.52} & \dvalp{\uline{90.17}}{0.40} & \dvalp{57.58}{2.98} \\
& Trace2Skill & \dvalp{44.93}{2.94} & \dvalp{\uline{69.21}}{3.21} & \dvalp{31.60}{4.95} & \dvalp{75.56}{9.52} & \dvaln{35.64}{1.52} & \dvaln{89.10}{0.67} & \dvalp{57.67}{3.07} \\
& SkillOpt-Lite & \dvalp{56.93}{14.94} & \dvalp{67.57}{1.57} & \dvalp{29.01}{2.36} & \dvalp{81.34}{15.30} & \dvalp{37.84}{0.68} & \dvalp{90.11}{0.34} & \dvalp{60.47}{5.87} \\
& SkillOpt & \dvalp{\uline{57.65}}{15.66} & \dvalp{67.36}{1.36} & \dvalp{\uline{69.81}}{43.16} & \dvalp{\uline{83.21}}{17.17} & \dvaln{34.63}{2.53} & \dvalp{89.77}{0.00} & \dvalp{\uline{67.07}}{12.47} \\
& \our & \dvalp{\textbf{61.48}}{19.49} & \dvalp{\textbf{71.29}}{5.29} & \dvalp{\textbf{71.46}}{44.81} & \dvalp{\textbf{86.38}}{20.34} & \dvalp{\textbf{40.03}}{2.87} & \dvalp{\textbf{91.24}}{1.47} & \dvalp{\textbf{70.31}}{15.71} \\
\bottomrule
\end{tabular}%
}
\vspace{-5pt}
\end{table*}

\subsection{Experimental Setup}
\label{sec:experimental_setup}

\paragraph{Datasets.}
The experiments are conducted on six benchmarks.
\emph{General question answering} tasks include DocVQA~\citep{mathew2021docvqa} and SearchQA~\citep{dunn2017searchqa}.
\emph{Mathematical reasoning} task is LiveMathematicianBench~(LiveMath; \citet{he2026livemathematicianbenchlivebenchmarkmathematicianlevel}).
\emph{Agentic tasks} include SpreadsheetBench~\citep{spreadsheetbench}, OfficeQA~\citep{opsahl2026officeqa}, and ALFWorld~\citep{alfworld}.
We adopt the data splits from SkillOpt-Lite~\citep{shen2026skillopt}.

\paragraph{Models.}
We employ GLM-5.3~\citep{zeng2026glm}, Qwen3.8-Flash-Next~\citep{qwen2026design}, DeepSeek-V4-Pro-0813~\citep{deepseekai2026deepseekv4}, DeepSeek-V4-Flash, and Qwen3.6-35B-A3B~\citep{qwen36_35b_a3b} as the target model.
For the first three models, we use themselves as the optimizer for a self-evolving setting.
For the less capable two models, DeepSeek-V4-Pro is employed as the optimizer to evaluate a setting where a stronger model guides a weaker model.

\paragraph{Settings.}
We compare our method against the following baselines: running without a skill, the initial skill written by human, skill-evolving methods Trace2Skil~\citep{ni2026trace2skill}, SkillOpt~\citep{yang2026skillopt}, and
SkillOpt-Lite~\citep{shen2026skillopt}.
And we follow the training protocol of SkillOpt-Lite.
The default number of rollouts $n$ is set to $8$.
Since all test sets other than SearchQA are on the scale of hundreds of samples, resulting in significant testing volatility, we evaluate these benchmarks over \textbf{four independent runs} and report the average accuracy.
More details are provided in Appendix~\ref{appendix:exp_details}.

\subsection{Main Results}
\label{sec:main_results}


As presented in Table~\ref{tab:skill_results}, \our achieves the highest scores across 26 model--benchmark pairs, outperforming baselines by a large margin.
Compared with the second-best method SkillOpt, our method further improves average accuracy by 5.69 points, demonstrating the benefits of learning from grouped trajectories.
On both the self-evolving setting and strong-guide-weak setting, \our exhibits consistent gains across LLMs from 35B to 1.6T, with average gains up to 27.23 over ``No skill''.
Notably, the OfficeQA performance of DeepSeek-V4-Flash improves from 5.41\% to 50.51\% with our method, surpassing the second-best method by 24.50\%.
This is associated with a recurring format error: the model frequently uses \texttt{\textless\textbar DSML\textbar answer\textgreater...\textless/\textbar DSML\textbar answer\textgreater} delimiters instead of the required \texttt{<answer>...</answer>} tags, preventing its responses from being recognized as valid answers.
\our retains this repeated pattern in memory and incorporates explicit instructions prohibiting DSML-style output into the skill.
This provides a concrete example of how cross-step memory supports the correction of persistent model-specific errors.
For more detailed analysis, please refer to Appendix~\ref{appendix:livemath}.

\subsection{Ablation Study}
\label{sec:ablation}

To investigate the impact of each component of \our, we conduct two ablations.
``SkillOpt w/ Memory'' augments the strongest baseline, SkillOpt, with the failure memory in our framework.
We also evaluate ``\our w/o Memory'', which removes dual memory while retaining group contrast analysis.
As shown in Table~\ref{tab:memory_ablation}, compared with SkillOpt, ``SkillOpt w/ Memory'' improves performance in 8 of the 11 model--benchmark pairs and reduces in the remaining three.
These mixed results suggest that adding memory to conventional single-rollout optimization yields gains, but not consistently.
In contrast, removing memory from \our decreases performance in all pairs.
Despite this, ``\our w/o Memory'' still exhibits improvement over Skillopt.
The contrasting effects of memory can be attributed to differences in the evidence within each update.
Under a comparable budget, single-rollout methods observe more distinct questions per step, potentially limiting the benefit of memory.
Grouped rollouts instead concentrate on repeated trajectories of fewer questions, making cross-step evidence particularly useful for calibrating the optimization direction.
Together, the ablations show that the two components of \our are effective and synergistic.

\begin{table*}[t]
\centering
\caption{
Ablation experiment of method components on two models.
}
\label{tab:memory_ablation}
\vspace{-5pt}
\setlength{\tabcolsep}{4pt}
\renewcommand{\arraystretch}{1.10}
\resizebox{\textwidth}{!}{%
\begin{tabular}{llccccccc}
\toprule
\textbf{Model} & \textbf{Method} & \textbf{Spreadsheet} & \textbf{SearchQA} & \textbf{LiveMath} & \textbf{ALFWorld} & \textbf{OfficeQA} & \textbf{DocVQA} & \textbf{Avg} \\
\midrule
\multirow{4}{*}{\shortstack[l]{DeepSeek-V4\\Flash}}
 & SkillOpt & 59.16 & \underline{71.57} & 68.63 & 74.44 & 26.01 & -- & 59.96 \\
 & SkillOpt w/ Memory & \underline{59.43} & 71.14 & \underline{68.86} & \underline{77.80} & \underline{47.80} & -- & 65.01 \\
 & \our w/o Memory & 57.92 & 71.43 & 63.92 & 75.37 & 37.84 & -- & \underline{61.30} \\
 & \our & \textbf{62.54} & \textbf{73.36} & \textbf{74.29} & \textbf{84.14} & \textbf{50.51} & -- & \textbf{68.97} \\
\midrule
\multirow{4}{*}{\shortstack[l]{Qwen3.6\\35B-A3B}}
 & SkillOpt & 57.65 & 67.36 & 69.81 & 83.21 & 34.63 & 89.77 & 67.07 \\
 & SkillOpt w/ Memory & 58.63 & \underline{69.07} & 67.92 & 81.16 & \underline{35.14} & \underline{89.91} & 66.97 \\
 & \our w/o Memory & \underline{58.72} & 68.57 & \underline{70.99} & \underline{83.40} & 34.46 & 89.84 & \underline{67.66} \\
 & \our & \textbf{61.48} & \textbf{71.29} & \textbf{71.46} & \textbf{86.38} & \textbf{40.03} & \textbf{91.24} & \textbf{70.31} \\
\bottomrule
\end{tabular}%
}
\vspace{-5pt}
\end{table*}

\section{Further Analysis}

\begin{figure*}[t!]
    \centering
    \includegraphics[width=\linewidth]{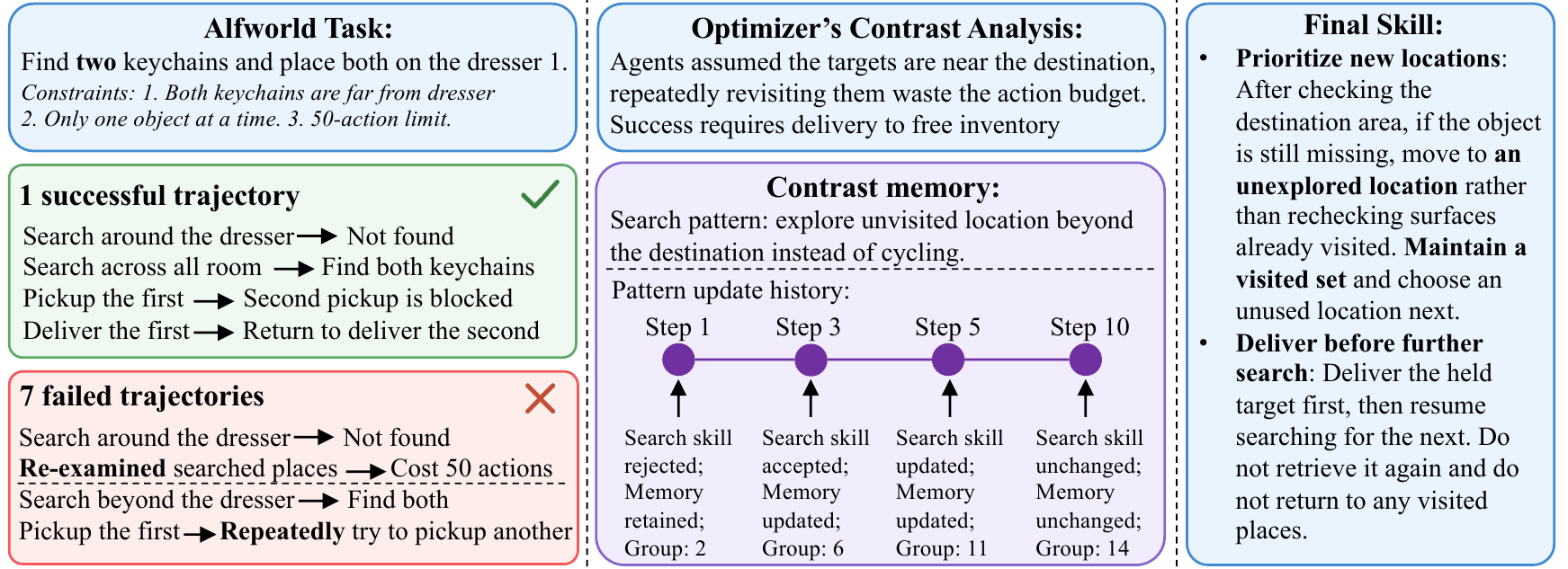}
    \vspace{-8pt}
    \caption{ 
    A case study of the evolution process of \our on ALFWorld, where the agent needs to manipulate objects within a simulated embodied environment.
    \our discovers a valuable pattern.
    Although the initial candidate skill is rejected, the extracted pattern remains in contrast memory.
    Its support group accumulates from 2 to 14 distinct tasks across optimization steps.
    During the process, the skill is accepted and refined multiple times.
    }
    \label{fig:case}
    \vspace{-5pt}
\end{figure*}


\subsection{Case Study}
\label{sec:case_study}

Figure~\ref{fig:case} showcases a detailed evolution process of \our.
As illustrated, group contrast reveals a useful search pattern: explore unvisited locations beyond the destination rather than repeatedly checking the same area.
With the accumulation of supporting groups in the contrast memory, the corresponding skill edit is selected and accepted at step 3.
At step 5, a further revision introduces a complementary rule: deliver the held target before searching for the next.
The resulting skill includes practical guidance absent from single-rollout baselines and leads to stronger test performance.


\subsection{Skill Transferability}
\label{sec:skill_transfer}

\begin{figure*}[t!]
    \centering
    \begin{subfigure}{0.38\linewidth}
        \centering
        \includegraphics[width=\linewidth]{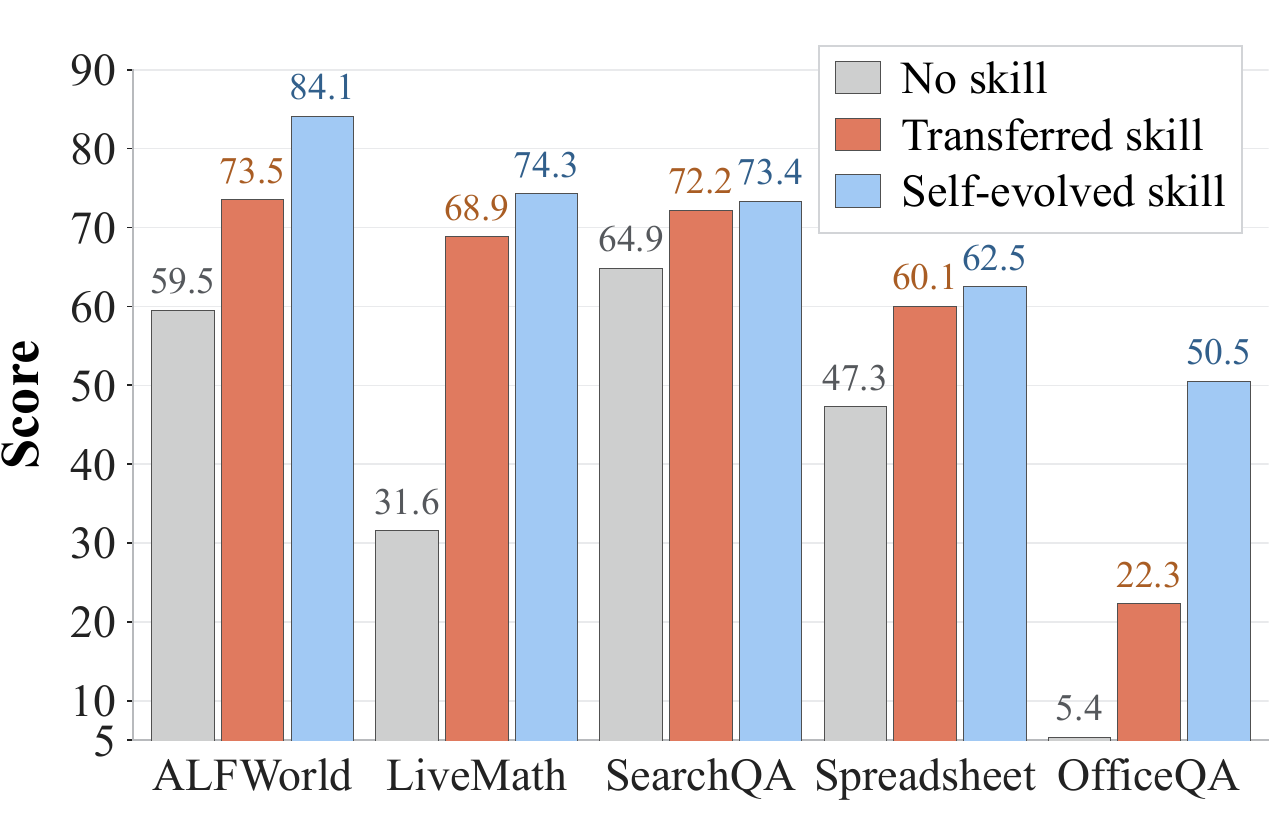}
        \caption{}
        \label{fig:cross-model}
    \end{subfigure}
    \begin{subfigure}{0.26\linewidth}
        \centering
        \includegraphics[width=\linewidth]{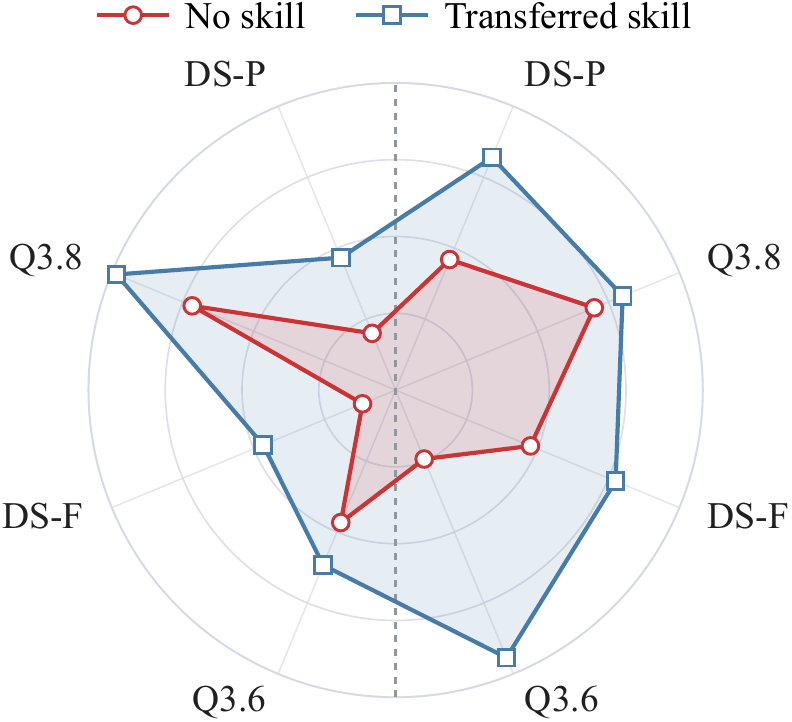}
        \caption{}
        \label{fig:cross-dataset}
    \end{subfigure}%
    \begin{subfigure}{0.34\linewidth}
        \centering
        \includegraphics[width=\linewidth]{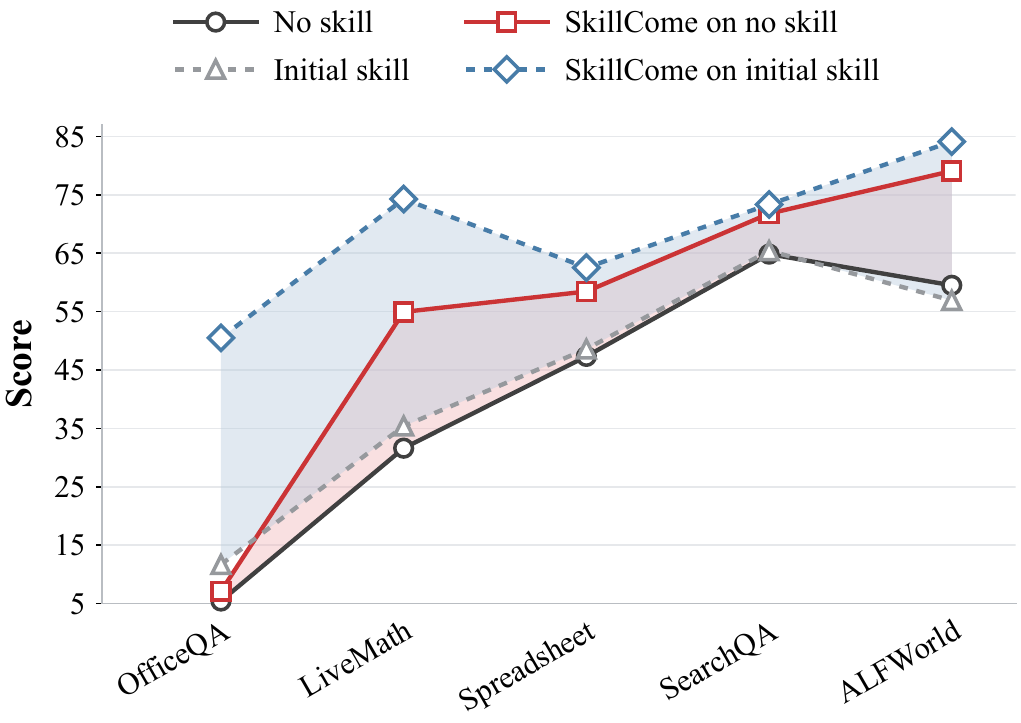}
        \caption{}
        \label{fig:evo_empty}
    \end{subfigure}
    \vspace{-5pt}
    \caption{
    (a) Cross-model skill transferability on DeepSeek-V4-Flash.
    (b) Cross-dataset skill transferability on four models (Names are abbreviated).
    Left are the HotpotQA results, while the right are the Omni-MATH results.
    (c) Evolution from empty skill on DeepSeek-V4-Flash.
    }
    \label{fig:analysis_three}
\vspace{-10pt}
\end{figure*}

\paragraph{Cross-Model Transferability.}
To evaluate whether the learned skills remain useful beyond the target model, we transfer the skills between DeepSeek-V4-Flash (Flash) and Qwen3.6-35B-A3B (Qwen3.6).
For each benchmark, a skill optimized for Qwen3.6 is directly applied to Flash.
Figure~\ref{fig:cross-model} shows that transferred skills significantly outperform the initial skill in all five benchmarks.
However, they still exhibit gaps from self-evolved skills, with the largest differences on OfficeQA.
This is consistent with previous discussion in \S\ref{sec:main_results}, since the skill learned on Qwen3.6 does not contain the DSML-style format issue.
Overall, these results indicate that our skills contain both broadly reusable procedures and corrections tailored to the behavioral patterns of individual models.

\paragraph{Cross-Dataset Transferability.}
We further evaluate the cross-dataset transferability of \our on four models.
Specifically, we directly transfer skills learned on SearchQA to HotpotQA~\citep{yang2018hotpotqa} and those learned on OlympiadBench~\citep{he2024olympiadbench} to Omni-MATH~\citep{gao2024omnimathuniversalolympiadlevel}.
As shown in Figure~\ref{fig:cross-dataset}, the transferred skills outperform the no-skill baseline on both benchmarks across all four models, demonstrating their generalizability to new datasets with similar task requirements.
More results on transferability are in Appendix~\ref{appendix:transfer_table} and ~\ref{appendix:cross_dataset}.

\subsection{Evolving from Empty Skill}
\label{sec:empty}

Since initial skills do not always improve performance, we investigate whether evolving from an empty skill provides a better alternative.
We compare evolving from no skill with evolving from initial skills on DeepSeek-V4-Flash.
As presented in Figure~\ref{fig:evo_empty}, evolution from an empty skill improves over ``No skill'' on all five benchmarks, but evolving from the initial skill achieves higher final performance.
This shows that \our can learn useful skills from scratch, while predefined guidance generally provides better starting points, even when their immediate benefits are limited.

\subsection{Evaluation in Agent Harnesses}
\label{sec:harness}

\begin{wrapfigure}{r}{0.35\textwidth}
    \vspace*{-15pt}
    \centering
    \includegraphics[width=0.98\linewidth]{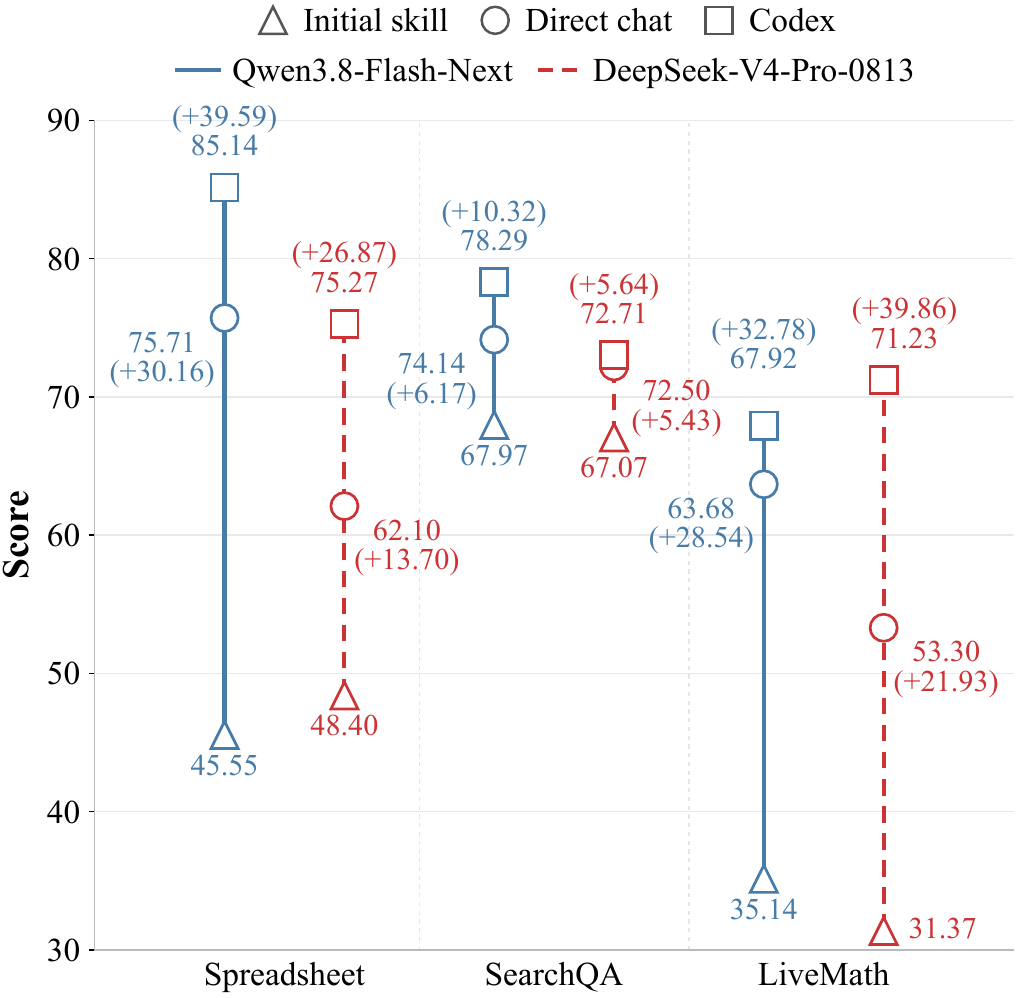}
    \vspace{-10pt}
    \caption{
    Performance comparison between Codex and Direct chat.
    }
    \vspace{-10pt}
    \label{fig:harness}
\end{wrapfigure}

To further evaluate our method in agent harnesses, we apply \our on Qwen3.8-Flash-Next and DeepSeek-V4-Pro-0813 within the Codex CLI harness on three datasets.
As illustrated in Figure~\ref{fig:harness}, we compare three settings: \emph{Initial skill}, where the agent operates through the chat completion mode under the initial skill; \emph{Direct chat}, where the agent operates through the chat completion mode under the skill learned by \our; and \emph{Codex}, where the agent operates within the Codex CLI harness under the skill learned by \our.
For both models, running under Codex consistently outperforms running under direct chat across all datasets.
These results demonstrate the effectiveness of our method within more advanced and complex harness environments, highlighting its practical value in real-world workflows.
\section{Conclusion}

In this paper, we identify that existing skill evolution methods lack not only analysis of different trajectories for the same question, but also historical information across training steps.
To mitigate these issues, we propose \our, a skill-evolution method based on group contrast analysis and a dual-memory system.
Combining the two synergistic components, \our derives precise optimization signals and generalizable optimization directions, learning effective and reusable skills.
Comprehensive evaluation across diverse models and benchmarks demonstrates that \our substantially improves agent performance, highlighting its potential for various real-world applications.

\section{AI use statement}

In this work, we used generative AI tools for proofreading and polishing the paper.
For example, we use LLMs to check the grammar and improve the readability.
We also used LLMs for literature discovery.
Specifically, they suggested lists of potentially relevant papers, which the authors then independently located the original publications.
The citations and descriptions of cited work were prepared by the authors.
All AI-generated content was manually reviewed by the authors.
We take responsibility for the final content of this work, including text, claims or artifacts produced with the aid of generative AI.

\section{Reproducibility Statement}
In this section, we list any related materials that help to reproduce this paper.. 
\begin{enumerate}[leftmargin=*]
    \item \textbf{Datasets}: The training and evaluation dataset we used is described in \S\ref{sec:experimental_setup}. The complete description of each dataset and the processing steps are provided in Appendix~\ref{appendix:dataset_details}.
    \item \textbf{Implementation Details}: The implementation details, such as training\&evaluation settings and hyperparameters are described in \S\ref{sec:experimental_setup}. And their details are provided in Appendix~\ref{appendix:implementation}.
    \item \textbf{Code}: The code to reproduce our algorithm is provided in the supplementary materials.
\end{enumerate}

\bibliography{iclr2027_conference}
\bibliographystyle{iclr2027_conference}

\clearpage
\onecolumn
\appendix 
\hypersetup{pdfborder={0 0 0}}
\etocdepthtag.toc{mtappendix}
\etocsettagdepth{mtchapter}{none}
\etocsettagdepth{mtappendix}{subsection}
\renewcommand{\contentsname}{Appendix}
\tableofcontents 
\clearpage
\hypersetup{pdfborder={0.5 0.5 0.5}}


\section{Method Details}
\label{appendix:method_details}

\begin{promptbox}[title={Prompt \thetcbcounter: Prompt for Group Contrast Reflection},label={prompt:same_task_contrast}]
\ttfamily
\fontsize{7}{8}\selectfont  
\raggedright
\setlength{\parindent}{2pt}
\setlength{\parskip}{2pt}

You are an expert same-task contrast analyst for AI agent trajectories.

\medskip
You receive several valid rollouts of the EXACT SAME task under the same
skill. At least one rollout succeeded and at least one failed. Analyze them
jointly; do not treat them as unrelated examples.

\medskip
\noindent Procedure:

\begin{enumerate}
    \setlength{\itemsep}{2pt}
    \setlength{\parsep}{0pt}

    \item Compare successful and failed trajectories and locate the earliest
    meaningful behavioral divergence.

    \item Explain what the successful rollout did that the failed rollout omitted,
    misunderstood, or executed incorrectly.

    \item Check whether the current skill already contains the needed guidance. If
    it does, sharpen it only when the failed behavior shows the rule is
    ambiguous or insufficiently actionable.

    \item Distinguish skill-relevant reasoning/strategy errors from random wording,
    evaluator noise, tool nondeterminism, and infrastructure behavior.

    \item Propose only generalizable guidance. Never encode the task id, its answer,
    entity names, file names, cell addresses, or other instance-specific facts.

    \item Prefer the smallest effective edit. It is correct to return no edits when
    the contrast does not support a causal, reusable lesson.

    \item Do not edit protected content between SLOW\_UPDATE markers.

    \item Compare the mechanism with the supplied cross-step hypothesis bank. Link
    an existing ID only when the causal mechanism is genuinely the same.
    Otherwise leave the related ID empty. A causal, generalizable contrast
    from one task remains eligible for an edit; cross-task support is ranking
    evidence, not an eligibility requirement.
\end{enumerate}

\noindent Respond ONLY with a valid JSON object:

\smallskip
{\ttfamily
\fontsize{7}{8}\selectfont  
\raggedright
\setlength{\parindent}{2pt}
\setlength{\parskip}{2pt}
\noindent\{\\
\hspace*{1em}"contrast\_summary": \{\\
\hspace*{2em}"first\_divergence": "\textless{}earliest important difference\textgreater{}",\\
\hspace*{2em}"success\_behavior": "\textless{}what worked\textgreater{}",\\
\hspace*{2em}"failure\_behavior": "\textless{}what went wrong\textgreater{}",\\
\hspace*{2em}"causal\_hypothesis": "\textless{}why this changed the outcome\textgreater{}",\\
\hspace*{2em}"generalizable": true,\\
\hspace*{2em}"confidence": 0.0\\
\hspace*{1em}\},\\
\hspace*{1em}"hypothesis": \{\\
\hspace*{2em}"related\_hypothesis\_id": "\textless{}existing bank ID or empty\textgreater{}",\\
\hspace*{2em}"mechanism": "\textless{}abstract reusable behavioral mechanism\textgreater{}"\\
\hspace*{1em}\},\\
\hspace*{1em}"patch": \{\\
\hspace*{2em}"reasoning": "\textless{}why these edits follow from the same-task contrast\textgreater{}",\\
\hspace*{2em}"edits": [\\
\hspace*{3em}\{\\
\hspace*{4em}"op": "append|insert\_after|replace|delete",\\
\hspace*{4em}"target": "\textless{}if needed\textgreater{}",\\
\hspace*{4em}"content": "\textless{}generalizable markdown guidance\textgreater{}"\\
\hspace*{3em}\}\\
\hspace*{2em}]\\
\hspace*{1em}\}\\
\}
\par
}
\end{promptbox}

In this section, we provide the prompt for the core component of \our, group contrast reflection.
As shown in Prompt~\ref{prompt:same_task_contrast}, the optimizer jointly analyzes successful and failed trajectories within the same group and returns a structured JSON response containing the contrast analysis, pattern information for contrast memory, and proposed skill edits.
Prompt~\ref{prompt:contrast_example} further presents an example output of group contrast reflection.
The example corresponds to the ALFWorld case discussed in Figure~\ref{fig:case} of \S\ref{sec:case_study}.
It is the output from training step 1, telling the agent to search the target beyond visited places around the destination.

\clearpage
\begin{promptbox}[breakable,title={Prompt \thetcbcounter: Example of Group Contrast Reflection Output},label={prompt:contrast_example}]
\begingroup
\ttfamily
\fontsize{7}{7}\selectfont  
\raggedright
\setlength{\parindent}{0pt}
\setlength{\parskip}{1pt}

\{ \par
\hspace*{1em}"contrast\_summary": \{ \par
\hspace*{2em}"first\_divergence": "The successful rollout eventually visited both sidetables after exhaustively searching dresser-adjacent locations, found the two keychains there, transported them one at a time to the dresser, and placed them. Most failed rollouts never visited the sidetables at all, or found the keychains late but then wandered back to already-searched receptacles instead of delivering them immediately.", \par

\hspace*{2em}"success\_behavior": "The successful agent treated the search as open-ended across all room furniture rather than only around the destination dresser. Once it found the first keychain, it handled the full inventory constraint by moving that keychain to the destination before returning for the second keychain, then delivered the second keychain to the same destination.", \par

\hspace*{2em}"failure\_behavior": "Failed trajectories repeatedly cycled through drawers and cabinets near the dresser, re-examined known-empty receptacles many times, and often never explored the sidetables containing the goal objects. Even trajectories that found one keychain frequently continued searching known-empty locations or failed to deliver held goal objects promptly.", \par

\hspace*{2em}"causal\_hypothesis": "Agents assumed the target objects would be near the destination and exhausted only its neighboring receptacles, repeatedly revisiting them instead of searching unexplored furniture elsewhere. This caused step-limit failure because the actual target objects were on unvisited sidetables. In addition, when an agent held one target object, it needed to place it before picking up the second, because inventory holds only one object.", \par

\hspace*{2em}"generalizable": true, \par
\hspace*{2em}"confidence": 0.82 \par
\hspace*{1em}\}, \par

\hspace*{1em}"hypothesis": \{ \par
\hspace*{2em}"related\_hypothesis\_id": "contrast\_v2\_cc5a59433591", \par
\hspace*{2em}"mechanism": "Agents fail multi-object search tasks when they limit exploration to receptacles near the stated destination and repeatedly revisit exhausted locations. Success requires maintaining unexplored-location coverage across the whole environment and, when inventory is capacity-limited and the task requires two objects, delivering the first target object before retrieving the second." \par
\hspace*{1em}\}, \par

\hspace*{1em}"patch": \{ \par
\hspace*{2em}"reasoning": "The same-task contrast shows failures not because the agents lacked the locate/take/place sequence but because they searched only near the destination and revisited known-empty receptacles. The existing skill already warns about revisiting searched locations, but it does not explicitly instruct agents to broaden exploration across the entire environment when local receptacles are empty, nor to deliver a held target object before attempting to pick up a second required object when inventory is limited. These are actionable, task-type-general extensions.", \par

\hspace*{2em}"edits": [ \par
\hspace*{3em}\{ \par
\hspace*{4em}"op": "append", \par
\hspace*{4em}"target": "Common Mistakes to Avoid", \par
\hspace*{4em}"content": "- **Destination-tunnel vision**: Do not search only receptacles adjacent to the final destination. If nearby places are empty, expand exploration to other furniture in the room before revisiting anything.\textbackslash{}n- **Holding one target while trying to take another**: If the task requires two objects and inventory is full, deliver the first held target to the destination before picking up the second.", \par
\hspace*{4em}"source\_type": "contrast", \par
\hspace*{4em}"support\_count": 2, \par
\hspace*{4em}"group\_support": 2, \par
\hspace*{4em}"trajectory\_support": 8, \par
\hspace*{4em}"memory\_refs": [ \par
\hspace*{5em}\{ \par
\hspace*{6em}"bank\_type": "contrast", \par
\hspace*{6em}"hypothesis\_id": "contrast\_v2\_cc5a59433591" \par
\hspace*{5em}\} \par
\hspace*{4em}], \par
\hspace*{4em}"hypothesis\_id": "contrast\_v2\_cc5a59433591" \par
\hspace*{3em}\} \par
\hspace*{2em}] \par
\hspace*{1em}\}, \par

\hspace*{1em}"source\_type": "contrast", \par
\hspace*{1em}"batch\_size": 8, \par
\hspace*{1em}"trajectory\_support": 8, \par
\hspace*{1em}"group\_support": 2, \par
\hspace*{1em}"group\_ids": [ \par
\hspace*{2em}"train/\allowbreak look\_at\_obj\_in\_light-\allowbreak Pen-None-DeskLamp-305/\allowbreak trial\_T20190907\_\allowbreak 115849\_734053", \par
\hspace*{2em}"train/\allowbreak pick\_clean\_then\_place\_in\_recep-\allowbreak Lettuce-None-GarbageCan-20/\allowbreak trial\_T20190909\_\allowbreak 033324\_286989", \par
\hspace*{2em}"train/\allowbreak pick\_two\_obj\_and\_place-\allowbreak KeyChain-None-Dresser-322/\allowbreak trial\_T20190908\_\allowbreak 051114\_312895" \par
\hspace*{1em}], \par
\hspace*{1em}"hypothesis\_id": "contrast\_v2\_cc5a59433591" \par
\}
\par
\endgroup
\end{promptbox}

\section{Detailed Experimental Setting}
\label{appendix:exp_details}

\subsection{Dataset Details}
\label{appendix:dataset_details}

\begin{table}[t]
\centering
\caption{The statistics of the datasets used in our paper.
Transfer-only target datasets have no training or validation split.}
\label{tab:dataset_statistics}
\setlength{\tabcolsep}{5pt}
\renewcommand{\arraystretch}{1.10}
\begin{tabular}{lrrrr}
\toprule
\textbf{Dataset} & \textbf{Train} & \textbf{Validation} &
\textbf{Test} & \textbf{Total} \\
\midrule
SearchQA         & 400 & 200 & 1400 & 2000 \\
SpreadsheetBench & 40  & 79  & 281  & 400  \\
OfficeQA         & 49  & 49  & 148  & 246  \\
DocVQA           & 107 & 53  & 374  & 534  \\
LiveMath         & 36  & 35  & 106  & 177  \\
ALFWorld         & 200 & 140 & 134  & 474  \\
\midrule
HotpotQA         & --  & --  & 1000 & 1000 \\
OlympiadBench    & 40  & 100 & 60   & 200  \\
Omni-MATH        & --  & --  & 1000 & 1000 \\
\bottomrule
\end{tabular}%
\end{table}

Table~\ref{tab:dataset_statistics} summarizes the dataset and splits adopted in our work.
The main evaluation covers six benchmarks spanning question answering, mathematical reasoning, and agentic tasks.
We mainly follow the splits from SkillOpt-Lite~\citep{shen2026skillopt}.
Detailed information on each dataset is as follows:

\begin{itemize}
    \item \textbf{SearchQA}~\citep{dunn2017searchqa} pairs trivia questions with supporting snippets retrieved from the web.
    Answering these questions requires extracting relevant facts from potentially noisy or redundant evidence.
    Here, models answer these questions using the provided snippets rather than performing live web searches.

    \item \textbf{SpreadsheetBench}~\citep{spreadsheetbench} evaluates spreadsheet manipulation through practical requests originating from real users.
    Agents must interpret natural-language instructions and implement Python code to perform the corresponding operations on spreadsheet files.

    \item \textbf{OfficeQA}~\citep{opsahl2026officeqa} evaluates question answering over lengthy financial documents containing narrative text and tables.
    We use the 246-question collection in an offline document-tool setting, where agents inspect locally available documents to produce their answers.

    \item \textbf{DocVQA}~\citep{mathew2021docvqa} evaluates visual question answering over document images.
    The model must read document content and interpret its spatial organization to locate the information requested by a question.

    \item \textbf{LiveMathematicianBench (LiveMath)}~\citep{he2026livemathematicianbenchlivebenchmarkmathematicianlevel} assesses research-level mathematical reasoning through multiple-choice questions constructed from mathematical research papers.

    \item \textbf{ALFWorld}~\citep{alfworld} evaluates sequential decision making in household environments through a text-based interaction interface.
    Agents must search for objects, manipulate them, and coordinate multiple actions to satisfy a specified goal.
\end{itemize}

We additionally use three datasets for cross-dataset transfer experiments.
The search-based benchmark HotpotQA~\citep{yang2018hotpotqa} is adopted to evaluate the skills evolved on SearchQA.
We also transfer the skills evolved on OlympiadBench~\citep{he2024olympiadbench} to Omni-MATH~\citep{gao2024omnimathuniversalolympiadlevel}.
OlympiadBench includes training, validation, and test splits, whereas HotpotQA and Omni-MATH are used exclusively for transfer evaluation without training or validation.

\begin{itemize}
    \item \textbf{HotpotQA}~\citep{yang2018hotpotqa} is a multi-hop question-answering benchmark based on Wikipedia articles.
    Its questions require combining evidence from multiple passages, including linking related facts and comparing information about different entities.
    We randomly select 1000 examples from its distractor validation split to evaluate the direct transfer of skills learned on SearchQA.

    \item \textbf{OlympiadBench}~\citep{he2024olympiadbench} contains challenging open-end mathematics and physics problems drawn from competitions and examinations.
    We use a mathematics subset as the source dataset for mathematical skill transfer.

    \item \textbf{Omni-MATH}~\citep{gao2024omnimathuniversalolympiadlevel} evaluates Olympiad-level mathematical problem solving across diverse mathematical topics and difficulty levels.
    We evaluate skills transferred from OlympiadBench on 1000 randomly selected samples on Omni-MATH.
\end{itemize}

\subsection{Implementation Details}
\label{appendix:implementation}

In this subsection, we provide additional implementation details.
Trace2Skill~\citep{ni2026trace2skill} uses its official implementation, which collects trajectories for one training epoch and analyzes them to construct the skill.
Our method uses the same training budget to that of SkillOpt.
Following the SkillOpt-Lite training protocol, we run SkillOpt and our method for four epochs or ten batches, whereas SkillOpt-Lite is executed for exactly ten batches.


For all models, the maximum length is set to 16384 tokens per response.
For grouped rollouts, we sample $n=8$ trajectories per question.
In practice, because group rollouts involve repeated rollouts for the same question, we deduplicate the input trajectories for failure and success reflection.
Since the test sets of the main benchmarks other than SearchQA contain only a few hundred examples, their scores can fluctuate across evaluations.
We therefore evaluate these benchmarks over four independent runs and report the average score to reduce the impact of stochastic variation.

\section{Additional Experiments}
\label{appendix:detailed_tables}

\subsection{Notable Findings in the Experiment}
\label{appendix:livemath}

\begin{figure}[th]
    \centering
    \begin{promptbox}[title={A LiveMath Question with a Meta-Option}]
    \small
    \textbf{Question (excerpt).}
    What is the strongest statement that can be proved about every sufficiently small perturbation $\psi_0\in\mathcal E$ of $\phi$ measured by $\rho_R$?

    \medskip
    \textbf{Option A.}
    One of the remaining options is correct, but a stronger result can be proven.

    \medskip
    \textbf{Options B--E (omitted).}
    Mathematical statements about stability with different conditions on the time domain, quantifiers, and phase or spatial shifts.

    \medskip
    \textbf{Ground-truth answer: A.}
    \end{promptbox}
    \caption{
    An example of the meta-option pattern in LiveMath.
    Option A is the meta-option and the ground-truth answer.
    The other options are summarized for brevity.
    }
    \label{fig:meta_option_example}
\end{figure}

\begin{figure*}[t]
\centering

\begin{tcolorbox}[
    title={(a) Meta-option Related Skill on DeepSeek-V4-Flash},
    colback=white,
    colframe=black!60,
    colbacktitle=black!60,
    coltitle=white,
    fonttitle=\bfseries\small,
    fontupper=\small,
    boxrule=0.6pt,
    boxsep=4pt,
    left=2pt,
    right=2pt,
    top=2pt,
    bottom=2pt,
    before skip=0pt,
    after skip=6pt
]
\textbf{Meta-Option Override (HARD RULE)}

\smallskip
\noindent
- If the question asks for the ``strongest,'' ``best possible,'' or
``maximal'' statement, OR if any option says ``One of the remaining
options is correct, but a stronger result can be proven'' (or its
equivalent), \textbf{STOP ALL ANALYSIS IMMEDIATELY}.
Do not read the explicit options. Do not compare hypotheses,
exponents, group structures, or any mathematical content.

\smallskip
\noindent
- \textbf{YOUR FINAL ANSWER IS THE META-OPTION.}
Output its letter and nothing else.
\end{tcolorbox}

\begin{tcolorbox}[
    title={(b) Meta-option Related Skill on GLM-5.3},
    colback=white,
    colframe=black!60,
    colbacktitle=black!60,
    coltitle=white,
    fonttitle=\bfseries\small,
    fontupper=\footnotesize,
    boxrule=0.6pt,
    boxsep=4pt,
    left=2pt,
    right=2pt,
    top=2pt,
    bottom=2pt,
    before skip=0pt,
    after skip=0pt
]
\textbf{Meta-Option Priority (UNCONDITIONAL DEFAULT)}

\smallskip
\noindent
When an option of the form ``One of the remaining options is correct,
but a stronger result can be proven'' is present, and the question
asks for the strongest provable statement, a complete
classification/characterization, the complete list, or ``which
description gives the complete...'', TAKE THE META-OPTION.
This is the single highest-yield rule: in the training history,
meta-options are correct far more often than any concrete option,
including on questions about spectral graph bounds, quantum K-theory,
von Neumann algebras, PDE regularity, commutator lengths, quasimodular
forms, and clique-sum decompositions.

\smallskip
\noindent
1. \textbf{CRITICAL --- a successful concrete derivation does NOT
license rejecting the meta-option.}
If you reconstruct a classical theorem (Schauder regularity, a spectral
extremal bound with equality case, a duality congruence, a structural
decomposition) and verify a concrete option matches it, that proves
ONLY truth, never maximality.
[...]

\smallskip
\noindent
2. \textbf{Only two legitimate exits from the meta-option:}
(a) you can name a specific, rigorous obstruction (rigidity theorem,
explicit counterexample to EVERY candidate strengthening, matching
sharp lower bound) valid under the FULL hypotheses; or
(b) the meta-option is internally contradicted.
Recall, derivation, confidence, and ``the concrete option already seems
sharp'' are NOT exits.

\smallskip
\noindent
[...]
\end{tcolorbox}
\vspace{10pt}
\caption{
Skills related to the Meta-option pattern on DeepSeek-V4-Flash and GLM-5.3.
The skill of DeepSeek-V4-Flash instructs the agent to select the meta-option without further analysis.
In contrast, the GLM-5.3 skill prioritizes the meta-option, but still requires the model to analyze the math problem itself.
[...] indicates omitted text.
}
\label{fig:meta_option_skills}
\end{figure*}

\begin{table}[th]
\centering
\caption{Livemath Performance with the Meta and Non-Meta subsets.
}
\label{tab:meta_nonmeta}
\setlength{\tabcolsep}{4pt}
\renewcommand{\arraystretch}{1.10}
\begin{tabular}{llccc}
\toprule
\textbf{Model} & \textbf{Skill} & \textbf{All} & \textbf{Meta} & \textbf{Non-Meta} \\
\midrule
\multirow{3}{*}{GLM-5.3}
 & No skill & 36.08 & 14.67 & 52.50 \\
\cmidrule(lr){2-5}
 & Init skill & 38.44 & 16.30 & 55.42 \\
 & SkillCome & \textbf{67.92} & \textbf{71.74} & \textbf{65.00} \\
\midrule
\multirow{3}{*}{\shortstack[l]{DeepSeek-V4\\Flash}}
 & No skill & 31.60 & 4.35 & 52.50 \\
\cmidrule(lr){2-5}
 & Init skill & 35.38 & 10.33 & 54.58 \\
 & SkillCome & \textbf{74.29} & \textbf{96.74} & \textbf{57.08} \\
\bottomrule
\end{tabular}%
\end{table}

As shown in Table~\ref{tab:skill_results} of \S\ref{sec:main_results}, the three more capable models, GLM-5.3, Qwen3.8-Flash-Next, and DeepSeek-V4-Pro-0813 do not always outperform DeepSeek-V4-Flash and Qwen3.6-35B-A3B after skill evolution, especially on LiveMath.
This is because some LiveMath questions include the option ``One of the remaining options is correct, but a stronger result can be proven.''
We refer to this as the \emph{meta-option}, since this option is always labeled as correct whenever it appears.
Figure~\ref{fig:meta_option_example} provides an example of questions with meta-option.
Of the 106 test samples, 46 contain this option and constitute the \emph{Meta} subset, while we refer to the remaining 60 as the \emph{Non-Meta} subset.
This regularity therefore provides a shortcut for answering the Meta subset without resolving the underlying mathematical problem.

The skills learned for DeepSeek-V4-Flash and Qwen3.6-35B-A3B adopt a strict rule for this pattern: once the meta-option is detected, the model should stop further reasoning and return its corresponding label.
By contrast, the skills learned for the stronger models also recognize the pattern but retain instructions to analyze the mathematical content before answering.
Figure~\ref{fig:meta_option_skills} contrasts the relevant skill learned for DeepSeek-V4-Flash and GLM-5.3.
These different strategies explain why the weaker target models achieve higher overall scores on LiveMath, since their gains mostly came from the Meta subset.

Table~\ref{tab:meta_nonmeta} reports the detailed results for each subset on DeepSeek-V4-Flash and GLM-5.3.
Relative to the initial skill, Flash improves from 10.33\% to 96.74\% on Meta, but only from 54.58\% to 57.08\% on Non-Meta.
GLM-5.3 exhibits a more balanced improvement, reaching 71.74\% on Meta and 65.00\% on Non-Meta.
Although Flash achieves a higher overall score, GLM-5.3 remains stronger on questions without the meta-option.
Overall, GLM-5.3 exhibits stronger mathematical reasoning capabilities.
The meta-option rule improves accuracy under the benchmark's labeling pattern, but its benefit should not be interpreted as an equivalent improvement in mathematical reasoning.

\subsection{Standard Deviation of Main Results}
\label{appendix:deviation_main}

\begin{table*}[ht]
\centering
\caption{
Standard deviations of \our.
Mean scores are provided in Table~\ref{tab:skill_results}.
``--'' denotes unsupported evaluation.
}
\label{tab:evaluation_std}
\setlength{\tabcolsep}{5pt}
\renewcommand{\arraystretch}{1.10}
\begin{tabular}{lccccc}
\toprule
\textbf{Model}
 & \textbf{Spreadsheet}
 & \textbf{LiveMath}
 & \textbf{ALFWorld}
 & \textbf{OfficeQA}
 & \textbf{DocVQA} \\
\midrule
Qwen3.8-Flash-Next
 & 1.78 & 1.81 & 1.83 & 2.09 & 0.69 \\
DeepSeek-V4-Pro-0813
 & 1.71 & 2.50 & 0.43 & 0.65 & -- \\
GLM-5.3
 & 0.92 & 2.83 & 0.43 & 1.50 & -- \\
DeepSeek-V4-Flash
 & 1.77 & 0.90 & 1.27 & 3.46 & -- \\
Qwen3.6-35B-A3B
 & 3.39 & 0.90 & 1.65 & 2.01 & 0.86 \\
\bottomrule
\end{tabular}%
\end{table*}

Table~\ref{tab:evaluation_std} reports the standard deviations of \our over four independent evaluation runs.
SearchQA is excluded because it is evaluated only once.
The relatively small standard deviations indicate that the performance of \our remains stable.

\subsection{Statistics of the Mix Group}
\label{appendix:mix_group}

\begin{table}[th]
\centering
\caption{
Mixed-group counts and proportions by target model.
A mixed group contains both successful and failed rollouts
of the same task.
Counts are reported as mixed groups / total groups.
}
\label{tab:mixed_groups_by_model}
\setlength{\tabcolsep}{5pt}
\renewcommand{\arraystretch}{1.10}
\begin{tabular}{lcc}
\toprule
\textbf{Target model} & \textbf{Mixed groups} & \textbf{Proportion} \\
\midrule
DeepSeek-V4-Flash    & 142 / 430  & 33.02\% \\
Qwen3.6-35B-A3B      & 192 / 480  & 40.00\% \\
GLM-5.3             & 107 / 430  & 24.88\% \\
DeepSeek-V4-Pro-0813 & 152 / 430  & 35.35\% \\
Qwen3.8-Flash-Next   & 138 / 480  & 28.75\% \\
\midrule
\textbf{Total} & \textbf{731 / 2250} & \textbf{32.49\%} \\
\bottomrule
\end{tabular}%
\end{table}

Table~\ref{tab:mixed_groups_by_model} summarizes the mixed groups collected during skill evolution.
The two multimodal LLMs, Qwen3.6-35B-A3B and Qwen3.8-Flash-Next, have more total group counts since their experiments also include DocVQA.
Across the five target models, 731 of the 2,250 groups contain both successful and failed trajectories.
The remaining non-mixed groups contain either only successful trajectories or only failed trajectories.
These statistics show that mixed groups account for a substantial fraction of the collected groups, supporting group contrast reflection across different models.


\subsection{Ablation on the Components of \our}
\label{appendix:ablation_table}

In this subsection, we further analyze our ablation experiment on the components of \our discussed in \S\ref{sec:ablation}.
Compared with SkillOpt, ``SkillOpt w/ Memory'' improves performance in 8 of the 11 model--benchmark pairs and reduces in the remaining six.
Although the gains are not consistent, it improves DeepSeek-V4-Flash on OfficeQA by 21.79 points over SkillOpt.
This can be attributed to the memory system, which helps to learn explicit guidance against the DSML-style answer format discussed in \S~\ref{sec:main_results}.

Meanwhile, ``\our w/o Memory'' generally shows better performance than Skillopt.
Together, the ablations demonstrate that both components of our method are effective.
More importantly, combining them leads to the best results on all model-benchmark pairs, suggesting that the two components are synergistic.
Group contrast enables deeper per-question analysis, while dual memory connects these findings with evidence from a broader range of samples.

\subsection{Evaluation in Agent Harnesses}
\label{appendix:harness}

\begin{table*}[ht]
\centering
\caption{
Performance comparison of SkillCome under direct chat and the Codex CLI harness.
}
\label{tab:codex_results}
\setlength{\tabcolsep}{4pt}
\renewcommand{\arraystretch}{1.10}
\begin{tabular}{llccc}
\toprule
\textbf{Model} & \textbf{Skill} & \textbf{Spreadsheet} & \textbf{SearchQA} & \textbf{LiveMath} \\
\midrule
\multirow{4}{*}{\shortstack[l]{Qwen3.8\\Flash-Next}}
 & No skill & 48.58 & 67.50 & 33.73 \\
\cmidrule(lr){2-5}
 & Init skill
 & \dvaln{45.55}{3.03}
 & \dvalp{67.97}{0.47}
 & \dvalp{35.14}{1.41} \\
 & SkillCome
 & \dvalp{\underline{75.71}}{27.13}
 & \dvalp{\underline{74.14}}{6.64}
 & \dvalp{\underline{63.68}}{29.95} \\
 & Codex-SkillCome
 & \dvalp{\textbf{85.14}}{36.56}
 & \dvalp{\textbf{78.29}}{10.79}
 & \dvalp{\textbf{67.92}}{34.19} \\
\cmidrule(lr){1-5}
\multirow{4}{*}{\shortstack[l]{DeepSeek-V4\\Pro-0813}}
 & No skill & 47.78 & 66.57 & 28.07 \\
\cmidrule(lr){2-5}
 & Init skill
 & \dvalp{48.40}{0.62}
 & \dvalp{67.07}{0.50}
 & \dvalp{31.37}{3.30} \\
 & SkillCome
 & \dvalp{\underline{62.10}}{14.32}
 & \dvalp{\underline{72.50}}{5.93}
 & \dvalp{\underline{53.30}}{25.23} \\
 & Codex-SkillCome
 & \dvalp{\textbf{75.27}}{27.49}
 & \dvalp{\textbf{72.71}}{6.14}
 & \dvalp{\textbf{71.23}}{43.16} \\
\bottomrule
\end{tabular}%
\end{table*}

The detailed results of our evaluation in agent harnesses are presented in Table~\ref{tab:codex_results}, corresponding to Figure~\ref{fig:harness} in \S\ref{sec:harness}.
For both models, running under Codex consistently outperforms running under direct chat across all datasets.
The performance gains from the harness environment and its tools are particularly pronounced on SpreadsheetBench and Livemath.
These results demonstrate the effectiveness of our method within advanced harness environments, highlighting its practical value in real-world workflows.

\subsection{Cross-Model Transferability}
\label{appendix:transfer_table}

\begin{table*}[th]
\centering
\caption{
Cross-target skill transferability experiments of \our.
``Qwen3.6 $\rightarrow$ Flash'' denotes transferring the skill learned on Qwen3.6-35B-A3B to DeepSeek-V4-Flash.
``Flash $\rightarrow$ Flash'' is the original performance of \our on the DeepSeek-V4-Flash.
}
\label{tab:cross_target_transfer}
\setlength{\tabcolsep}{4pt}
\renewcommand{\arraystretch}{1.10}
\resizebox{\textwidth}{!}{%
\begin{tabular}{llcccccc}
\toprule
\textbf{Target Model} & \textbf{Skill} & \textbf{Spreadsheet} & \textbf{SearchQA} & \textbf{LiveMath} & \textbf{ALFWorld} & \textbf{OfficeQA} & \textbf{Avg} \\
\midrule
\multirow{4}{*}{\shortstack[l]{DeepSeek-V4\\Flash}}
& No skill & 47.33 & 64.85 & 31.60 & 59.51 & 5.41 & 41.74 \\
\cmidrule(lr){2-8}
& Init skill & \dvalp{48.58}{1.25} & \dvalp{65.43}{0.58} & \dvalp{35.38}{3.78} & \dvaln{56.91}{2.60} & \dvalp{11.66}{6.25} & \dvalp{43.59}{1.85} \\
& Qwen3.6 $\rightarrow$ Flash & \dvalp{60.05}{12.72} & \dvalp{72.21}{7.36} & \dvalp{68.87}{37.27} & \dvalp{73.51}{14.00} & \dvalp{22.30}{16.89} & \dvalp{59.39}{17.65} \\
& Flash $\rightarrow$ Flash & \dvalp{\textbf{62.54}}{15.21} & \dvalp{\textbf{73.36}}{8.51} & \dvalp{\textbf{74.29}}{42.69} & \dvalp{\textbf{84.14}}{24.63} & \dvalp{\textbf{50.51}}{45.10} & \dvalp{\textbf{68.97}}{27.23} \\
\midrule
\multirow{4}{*}{\shortstack[l]{Qwen3.6\\35B-A3B}}
& No skill & 41.99 & 66.00 & 26.65 & 66.04 & 37.16 & 47.57 \\
\cmidrule(lr){2-8}
& Init skill & \dvaln{40.75}{1.24} & \dvalp{66.23}{0.23} & \dvalp{30.19}{3.54} & \dvalp{79.48}{13.44} & \dvalp{38.68}{1.52} & \dvalp{51.07}{3.50} \\
& Flash $\rightarrow$ Qwen3.6 & \dvalp{60.23}{18.24} & \dvalp{70.79}{4.79} & \dvalp{\textbf{72.17}}{45.52} & \dvalp{83.02}{16.98} & \dvalp{\textbf{40.71}}{3.55} & \dvalp{65.38}{17.81} \\
& Qwen3.6 $\rightarrow$ Qwen3.6 & \dvalp{\textbf{61.48}}{19.49} & \dvalp{\textbf{71.29}}{5.29} & \dvalp{71.46}{44.81} & \dvalp{\textbf{86.38}}{20.34} & \dvalp{40.03}{2.87} & \dvalp{\textbf{66.13}}{18.56} \\
\bottomrule
\end{tabular}
}
\end{table*}

Figure~\ref{fig:cross-model} in \S~\ref{sec:skill_transfer} presents the transfer of skills learned on Qwen3.6-35B-A3B to DeepSeek-V4-Flash.
We also have evaluated the reverse direction and report the complete results in Table~\ref{tab:cross_target_transfer}.
Skills learned on DeepSeek-V4-Flash transfer well to Qwen3.6-35B-A3B, even outperforming skills evolved specifically for the target model on LiveMath and OfficeQA.
Conversely, as discussed in \S~\ref{sec:skill_transfer}, the transferred skills on OfficeQA of DeepSeek-V4-Flash shows substantially lower OfficeQA performance than self-evolved skills, because the transferred skill does not explicitly address Flash's DSML-style answer format.
These results indicate that \our learns not only reusable task-level strategies that generalize across different models, but also model-specific rules that correct distinctive behavioral errors.

\subsection{Cross-Dataset Transferability}
\label{appendix:cross_dataset}

\begin{table*}[th]
\centering
\caption{
Full results of cross-dataset skill transferability. $\Delta$ denotes the score improvement of transferred skill over no skill.
}
\label{tab:dataset_transfer}
\setlength{\tabcolsep}{4pt}
\renewcommand{\arraystretch}{1.10}
\begin{tabular}{lllccc}
\toprule
\textbf{Source dataset} & \textbf{Target dataset} & \textbf{Model} & \textbf{No skill} & \textbf{Transferred skill} & $\boldsymbol{\Delta}$ \\
\midrule
\multirow{4}{*}{SearchQA}
 & \multirow{4}{*}{HotpotQA}
 & DeepSeek-V4-Pro-0813 & 63.20 & 64.80 & +1.60 \\
 & & Qwen3.8-Flash-Next & 66.30 & 67.90 & +1.60 \\
 & & DeepSeek-V4-Flash & 62.70 & 64.80 & +2.10 \\
 & & Qwen3.6-35B-A3B & 64.80 & 65.70 & +0.90 \\
\midrule
\multirow{4}{*}{OlympiadBench}
 & \multirow{4}{*}{Omni-MATH}
 & DeepSeek-V4-Pro-0813 & 72.30 & 74.10 & +1.80 \\
 & & Qwen3.8-Flash-Next & 86.40 & 86.60 & +0.20 \\
 & & DeepSeek-V4-Flash & 58.90 & 60.10 & +1.20 \\
 & & Qwen3.6-35B-A3B & 66.70 & 71.60 & +4.90 \\
\bottomrule
\end{tabular}%
\end{table*}

Table~\ref{tab:dataset_transfer} reports the complete results corresponding to Figure~\ref{fig:cross-dataset} in \S~\ref{sec:skill_transfer}.
We transfer the skills learned on SearchQA to HotpotQA, a similar search-based task.
Following SkillOpt, we also transfer skills evolved from OlympiadBench to Omni-MATH, another benchmark for open-ended mathematical reasoning.
The transferred skills improve performance across all eight model--benchmark pairs, with gains ranging from 0.2 to 4.9 points.
Although the magnitude of improvement varies across models, the consistent gains on both target datasets indicate that the learned skills provide reusable guidance beyond the datasets on which they were optimized.
These results support the cross-dataset generalizability of \our within similar domains.

\subsection{Evolving from Empty Skill}
\label{appendix:empty_skill}

\begin{table*}[th]
\centering
\caption{
Performance of \our evolving from empty skills.
}
\label{tab:initialization_ablation}
\setlength{\tabcolsep}{4pt}
\renewcommand{\arraystretch}{1.10}
\resizebox{\textwidth}{!}{%
\begin{tabular}{llccccccc}
\toprule
\textbf{Model} & \textbf{Skill} & \textbf{Spreadsheet} & \textbf{SearchQA} & \textbf{LiveMath} & \textbf{ALFWorld} & \textbf{OfficeQA} & \textbf{DocVQA} & \textbf{Avg} \\
\midrule
\multirow{4}{*}{\shortstack[l]{DeepSeek-V4\\Flash}}
 & No skill & 47.33 & 64.85 & 31.60 & 59.51 & 5.41 & -- & 41.74 \\
 & \our
 & \dvalp{\underline{58.45}}{11.12}
 & \dvalp{\underline{71.79}}{6.94}
 & \dvalp{\underline{54.95}}{23.35}
 & \dvalp{\underline{79.10}}{19.59}
 & \dvalp{7.09}{1.68}
 & --
 & \dvalp{\underline{54.28}}{12.54} \\
\cmidrule(lr){2-9}
 & Init skill & 48.58 & 65.43 & 35.38 & 56.91 & \underline{11.66} & -- & 43.59 \\
 & \our
 & \dvalp{\textbf{62.54}}{13.96}
 & \dvalp{\textbf{73.36}}{7.93}
 & \dvalp{\textbf{74.29}}{38.91}
 & \dvalp{\textbf{84.14}}{27.23}
 & \dvalp{\textbf{50.51}}{38.85}
 & --
 & \dvalp{\textbf{68.97}}{25.38} \\
\midrule
\multirow{4}{*}{\shortstack[l]{Qwen3.6\\35B-A3B}}
 & No skill & 41.99 & 66.00 & 26.65 & 66.04 & 37.16 & 89.77 & 54.60 \\
 & \our
 & \dvalp{\underline{58.01}}{16.02}
 & \dvalp{\underline{70.07}}{4.07}
 & \dvalp{\underline{64.86}}{38.21}
 & \dvalp{\textbf{87.13}}{21.09}
 & \dvaln{33.28}{3.88}
 & \dvalp{90.04}{0.27}
 & \dvalp{\underline{67.23}}{12.63} \\
\cmidrule(lr){2-9}
 & Init skill & 40.75 & 66.23 & 30.19 & 79.48 & \underline{38.68} & \underline{90.17} & 57.58 \\
 & \our
 & \dvalp{\textbf{61.48}}{20.73}
 & \dvalp{\textbf{71.29}}{5.06}
 & \dvalp{\textbf{71.46}}{41.27}
 & \dvalp{\underline{86.38}}{6.90}
 & \dvalp{\textbf{40.03}}{1.35}
 & \dvalp{\textbf{91.24}}{1.07}
 & \dvalp{\textbf{70.31}}{12.73} \\
\bottomrule
\end{tabular}%
}
\end{table*}

As shown in the main Table~\ref{tab:skill_results}, initial skills do not consistently improve performance over ``No skill''.
For example, both Qwen models perform worse with the initial skill on Spreadsheet, while several other model--benchmark pairs exhibit only marginal gains.
These observations raise the question of whether skill evolution can achieve better results when starting without an initial skill.
We therefore investigate the setting of evolving from empty skill on DeepSeek-V4-Flash in Figure~\ref{fig:evo_empty} of \S\ref{sec:empty}.
Here, we provide more result on Qwen3.6-35B-A3B.

The full results are presented in Table~\ref{tab:initialization_ablation}.
Skill evolution from an empty skill improves over the ``No skill'' baseline in 10 of the 11 model--benchmark pairs, demonstrating that \our can learn useful guidance without a predefined skill.
Nevertheless, evolution from the initial skill achieves higher final accuracy in 10 of the 11 pairs.
The only exception is ALFWorld on Qwen3.6-35B-A3B, where starting from an empty skill yields a slight advantage.
Notably, on Spreadsheet with Qwen3.6-35B-A3B, the initial skill reduces performance before evolution but leads to a better final result, suggesting that a skill's immediate effectiveness does not fully reflect its value for evolution.

One explanation is that an initial skill provides task-specific procedures and constraints that offer a useful starting point for revision, even when some instructions are ineffective.
Under a limited optimization budget, the optimizer model can refine this existing guidance and correct problematic rules, rather than construct and organize the skill from scratch.
Such guidance may also influence the trajectories collected during evolution, providing more informative behavioral references for later updates.
Overall, these results suggest that skill initialization affects not only the starting performance but also the subsequent evolution process.
Although an empty initialization remains a viable alternative, an initial skill generally provides a more effective basis for evolution.

\subsection{Token Usage of Optimizer Model}
\label{appendix:token}

\begin{table*}[th]
\centering
\caption{
Optimizer token usage in millions (M).
The ratio is computed as \our / SkillOpt.
}
\label{tab:optimizer_token_cost}
\setlength{\tabcolsep}{5pt}
\renewcommand{\arraystretch}{1.10}
\begin{tabular}{llccc}
\toprule
\textbf{Target model} & \textbf{Dataset}
 & \textbf{\our{} (M)} & \textbf{SkillOpt (M)}
 & \textbf{Ratio} \\
\midrule
\multirow{5}{*}{\shortstack[l]{DeepSeek-V4\\Flash}}
 & SpreadsheetBench & 0.69 & 1.50 & $0.460\times$ \\
 & SearchQA         & 1.69 & 4.64 & $0.364\times$ \\
 & OfficeQA         & 0.97 & 2.20 & $0.441\times$ \\
 & LiveMath         & 0.56 & 0.90 & $0.622\times$ \\
 & ALFWorld         & 1.09 & 1.57 & $0.694\times$ \\
\midrule
\multirow{6}{*}{\shortstack[l]{Qwen3.6\\35B-A3B}}
 & SpreadsheetBench & 0.56 & 1.31 & $0.427\times$ \\
 & SearchQA         & 2.15 & 5.07 & $0.424\times$ \\
 & OfficeQA         & 0.75 & 1.86 & $0.403\times$ \\
 & DocVQA           & 0.16 & 0.45 & $0.356\times$ \\
 & LiveMath         & 0.39 & 1.06 & $0.368\times$ \\
 & ALFWorld         & 1.05 & 1.67 & $0.629\times$ \\
\bottomrule
\end{tabular}%
\end{table*}

Since the target model rollout budget is matched during training, we compare the optimizer model token usage of \our and SkillOpt.
As shown in Table~\ref{tab:optimizer_token_cost}, \our consumes fewer optimizer tokens across all evaluated datasets for both target models.
This reduction primarily comes from lower input token usage through trajectory deduplication under grouped rollouts, as described in Appendix~\ref{appendix:implementation}.
These results demonstrate that \our achieves substantial performance gains while reducing optimizer token costs.

\end{document}